%% file: example_paper.tex
\documentclass{article}

\usepackage{microtype}
\usepackage{graphicx}
\usepackage{subfigure}
\usepackage{booktabs} 
\usepackage{caption}
\usepackage{wrapfig}
\usepackage{graphicx}
\usepackage{adjustbox}
\usepackage{subfigure}
\usepackage[dvipsnames]{xcolor}
\usepackage{url}  
\usepackage[breaklinks=true]{hyperref}  

\usepackage{amsmath}
\usepackage{amssymb}
\usepackage{mathtools}
\usepackage{amsthm}
\usepackage{caption}
\usepackage{enumitem}

\usepackage{multirow}
\usepackage{multicol}

\theoremstyle{plain}

\theoremstyle{definition}

\theoremstyle{remark}

\DeclareMathOperator*{\argminA}{arg\,min} 

\usepackage[accepted]{mlsys2025}

\mlsystitlerunning{Self-Data Distillation for Recovering Quality in Pruned Large Language Models}

\begin{document}

\twocolumn[
\mlsystitle{Self-Data Distillation for Recovering Quality in \\ Pruned Large Language Models}




\begin{mlsysauthorlist}
\mlsysauthor{Vithursan Thangarasa}{to}
\mlsysauthor{Ganesh Venkatesh}{to}
\mlsysauthor{Mike Lasby}{calgary}\textsuperscript{\hspace{-3pt}*}\hspace{3pt}
\mlsysauthor{Nish Sinnadurai}{to}
\mlsysauthor{Sean Lie}{to} \\
\end{mlsysauthorlist}

\mlsysaffiliation{to}{Cerebras Systems, Sunnyvale, California}
\mlsysaffiliation{calgary}{University of Calgary, Calgary, Alberta. \textsuperscript{*}Work completed while on internship at Cerebras}

\mlsyscorrespondingauthor{Vithursan Thangarasa}{vithu@cerebras.net}

\mlsyskeywords{Machine Learning, MLSys}

\vskip 0.3in

\input{tex_files/abstract.tex}
]
\printAffiliationsAndNotice{}

\input{tex_files/introduction.tex}
\input{tex_files/method.tex}
\input{tex_files/ablation.tex}
\input{tex_files/empirical_results.tex}
\input{tex_files/speculative_decode.tex}
\input{tex_files/related_work.tex}
\input{tex_files/conclusion.tex}

\input{tex_files/acknowledgements.tex}
\bibliography{refs_nourl}
\bibliographystyle{mlsys2025}

\clearpage
\newpage
\appendix

\input{tex_files/appendix.tex}



\end{document}

%% file: tex_files/abstract.tex
\begin{abstract} 
Large language models have driven significant progress in natural language
processing, but their deployment requires substantial compute and memory
resources. As models scale, compression techniques become essential for
balancing model quality with computational efficiency. Structured pruning, which
removes less critical components of the model, is a promising strategy for
reducing complexity. However, one-shot pruning often results in significant
quality degradation, particularly in tasks requiring multi-step reasoning. To
recover lost quality, supervised fine-tuning (SFT) is commonly applied, but it
can lead to catastrophic forgetting by shifting the model's learned data
distribution. Therefore, addressing the degradation from both pruning and SFT is
essential to preserve the original model's quality. In this work, we utilize
\textit{self-data distilled fine-tuning} to address these challenges. Our
approach leverages the original, unpruned model to generate a distilled dataset
that preserves semantic richness and mitigates catastrophic forgetting by
maintaining alignment with the base model's knowledge. Empirically, we
demonstrate that self-data distillation consistently outperforms standard SFT,
improving average accuracy by up to 8\% on the HuggingFace OpenLLM Leaderboard
v1. Specifically, when pruning six decoder blocks on Llama3.1-8B Instruct (i.e.,
32 to 26 layers, reducing the model size from 8.03B to 6.72B parameters), our
method retains 91.2\% of the original model's accuracy compared to 81.7\% with
SFT, while reducing real-world FLOPs by 16.3\%. Furthermore, combining
self-data distilled models through model merging yields enhanced quality
retention. Additionally, leveraging these pruned models in speculative decoding
increases token acceptance rates, thereby improving inference efficiency in
applied settings.
\end{abstract}

%% file: tex_files/introduction.tex
\vspace{-5pt}
\section{Introduction}
The advent of large language models (LLMs) such as
GPT-4~\citep{openai2024gpt4technicalreport},
Gemini~\citep{geminiteam2024geminifamilyhighlycapable}, and Llama
3~\citep{dubey2024llama3herdmodels} has revolutionized natural language
processing (NLP), driving significant advancements across various tasks through
extensive pre-training on textual data. These models, enhanced by supervised
fine-tuning (SFT), demonstrate impressive instruction-following
abilities~\citep{ouyang2022traininglanguagemodelsfollow,
touvron2023llamaopenefficientfoundation}, but come with high compute costs for
both training and inference~\citep{kaplan2020scalinglawsneurallanguage,
hoffmann2022trainingcomputeoptimallargelanguage}. To address diverse deployment
requirements across varying model scales, sizes, and compute budgets,
compressing models for efficient inference is essential, particularly given the
significant time, data, and resource constraints associated with training
multiple multi-billion parameter models from scratch.

Most model compression techniques can be grouped into four main categories:
knowledge distillation (KD)~\citep{hinton2015distilling},
factorization~\citep{hu2022lora}, pruning~\citep{lecunBrain1989}, and
quantization~\citep{han2015deep}. In our work, we focus on pruning, though we
aim for our method to inspire further developments across these other
compression methods. Structured pruning, which selectively removes less critical
components of a neural network, has emerged as a promising method for improving
LLM efficiency~\citep{ma2023llmpruner}. This method has gained attention for its
ability to reduce memory and compute requirements, making inference more
efficient. Recent works have shown that LLMs exhibit significant redundancy,
particularly in the middle layers, where removing these layers has a minimal
impact on overall model
quality~\citep{men2024shortgptlayerslargelanguage,gromov2024unreasonableineffectivenessdeeperlayers}.
The residual stream of the Transformer~\citep{waswaniTransformers2017}
architecture is only slightly modified by the output of non-essential layers,
enabling the removal of these layers without drastically harming model quality. 

\begin{figure}[ht]
    \centering
    \includegraphics[width=\columnwidth]{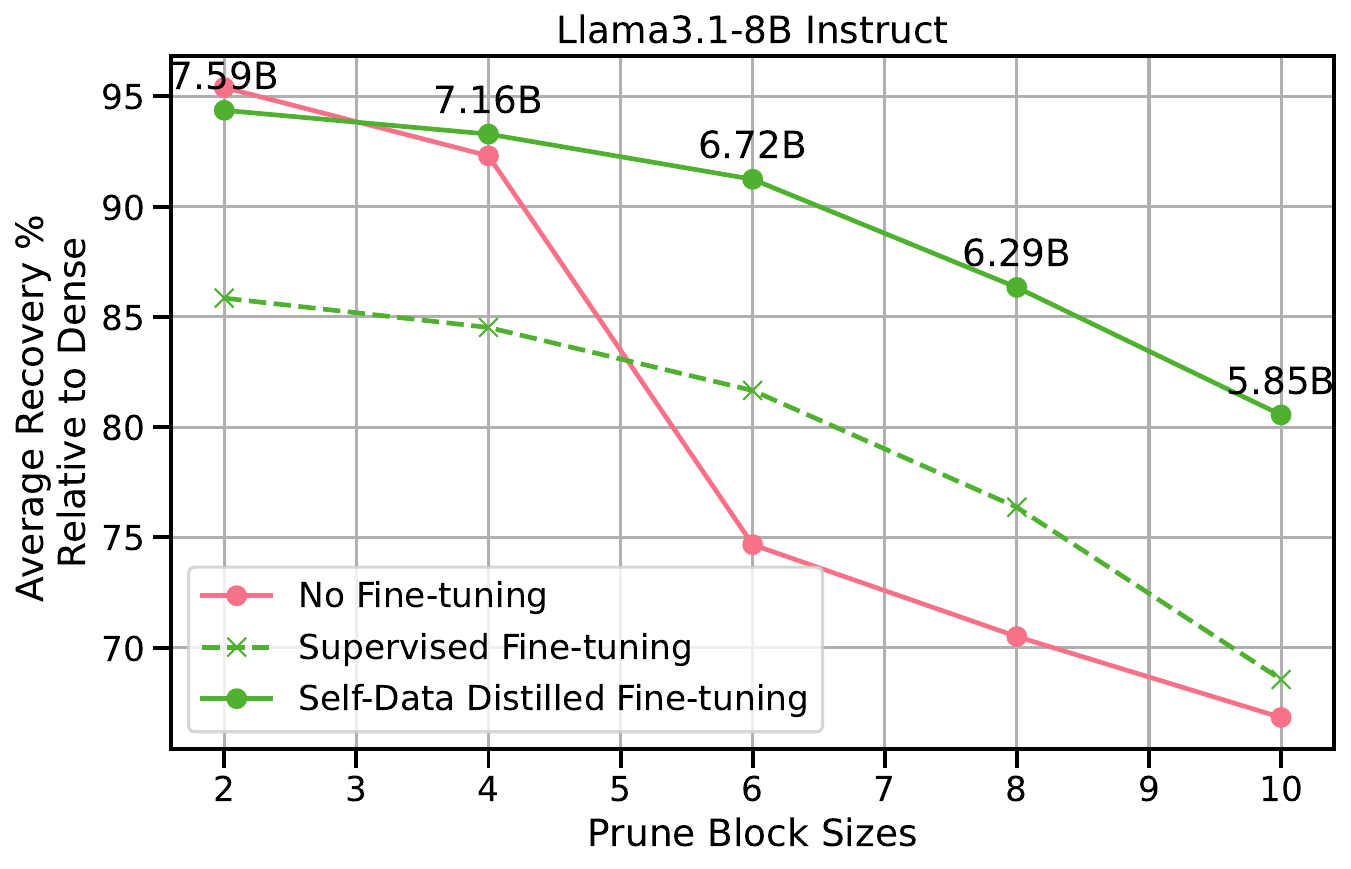}
    \caption{\textbf{Average quality recovery (\%) of pruned Llama3.1-8B
    Instruct models relative to the unpruned baseline, across varying prune
    block sizes on the HuggingFace OpenLLM Leaderboard v1.} The plot compares no
    fine-tuning, supervised fine-tuning, and self-data distilled fine-tuning
    using the OpenMathInstruct dataset. While model quality declines with prune
    block sizes, self-data distillation consistently achieves superior
    recovery.}
    \label{fig:recovery}
    \vspace{-5pt}
\end{figure}

Despite its potential advantages, depth-wise structured pruning presents
inherent challenges. It often leads to accuracy degradation, especially on tasks
requiring multi-step reasoning, such as
ARC-C~\citep{clark2018thinksolvedquestionanswering} or
GSM8k~\citep{cobbe2021trainingverifierssolvemath}, where the structured order of
layer outputs plays a crucial role. In these cases, pruning disrupts the flow of
information between layers, resulting in poor model quality even after
supervised fine-tuning (SFT)~\citep{sun2024transformerlayerspainters}. While SFT
can help recover some of the lost quality, it is generally insufficient for
tasks with high reasoning complexity, where the structured sequence of layer
outputs is essential. In addition, fine-tuning can amplify catastrophic
forgetting~\citep{mcclskey1989,kotha2024understanding}, where the model loses
previously learned information, particularly on tasks not represented in the
fine-tuning data. Standard mitigation strategies, such as data
replay~\citep{Ostapenko2022ContinualLW} or parameter importance-based
methods~\citep{KirkpatrickEWC2017}, often become impractical for LLMs due to
their scale. Moreover, fine-tuning often leads to distribution shifts, further
degrading model quality~\citep{Yang2024SelfDistillationBD}. As LLMs continue to
grow in size and complexity, developing more effective strategies to mitigate
these challenges during pruning is critical to unlocking its full potential.

In our work, we propose a novel approach to mitigate the adverse effects of
structured pruning by employing \textit{self-data distilled fine-tuning}. Our
method leverages the original, unpruned model as a seed language model to
generate a distilled dataset that upholds semantic equivalence with the original
task dataset. This approach not only preserves the semantic richness of the data
but also mitigates catastrophic forgetting, a phenomenon where fine-tuned models
lose their general instruction-following abilities due to the distribution shift
introduced during standard SFT. As seen in Figure~\ref{fig:recovery}, we show
that self-data distillation improves accuracy recovery by greater than 10\% over
SFT on the HuggingFace OpenLLM Leaderboard v1~\citep{open-llm-leaderboard}.
Specifically, when pruning 6 blocks from Llama3.1-8B Instruct, our approach
retains 91.2\% of the original model's accuracy compared to 81.7\% with SFT,
while also reducing FLOPs by 16.30\%. In addition to improving the quality of
pruned models, our work explores extending self-data distilled fine-tuning to
speculative decoding~\citep{leviation2023,
chen2023acceleratinglargelanguagemodel}. This approach allows pruned models to
not only recover their accuracy but also achieve efficient inference through
parallel token generation. The integration of self-data distillation with
speculative decoding is a natural extension, as it leverages the semantic
alignment achieved during data distillation to enhance the speculative
capabilities of pruned models. By aligning the pruned model's predictions more
closely with the target model, speculative decoding can benefit from improved
token acceptance rates, reducing latency while preserving model quality. The
main contributions of our work are:

\begin{itemize}
    \item To our knowledge, we are the first to introduce self-data distillation
    as a fine-tuning method for recovering the model quality of pruned models.
    Empirically, we show that self-data distillation on Llama3.1-8B Instruct and
    Mistral-7B-v0.3 Instruct~\citep{jiang2023mistral7b} consistently outperforms
    SFT across all pruned models.
    \item We demonstrate that self-data distillation scales effectively across a
    wide range of open-source fine-tuning datasets for LLMs, covering
    open-domain conversation, reasoning, and instruction following, with quality
    recovery significantly improving as the dataset size increases.
    \item We extend self-data distilled fine-tuning to speculative decoding,
    showing that it enhances token acceptance rates and reduces latency during
    inference. This extension highlights the broader applicability of our
    approach, supporting both compression and acceleration of large language
    models.
\end{itemize}

%% file: tex_files/method.tex
\vspace{-5pt}
\section{Methodology}
In this section, we present our approach to enhancing the efficiency of LLMs
through \textit{structured layer pruning} combined with \textit{self-data
distillation}. Our strategy involves systematically identifying and removing
redundant layers to optimize model efficiency while preserving task-specific
accuracy. Post-pruning, we employ self-data distillation to mitigate the effects
of catastrophic forgetting during the fine-tuning phase, thereby ensuring that
the pruned model improves its quality over standard SFT.

\vspace{-5pt}
\subsection{Layer-Pruning Algorithm for Language Models}
\label{subsec:layer-pruning-algo}

Transformers~\citep{waswaniTransformers2017} have become a foundational
architecture in deep learning, particularly for tasks involving natural language
processing and sequence modeling. A standard Transformer consists of \(L\)
layers, each of which includes a multi-head self-attention mechanism followed by
a feedforward network. These layers sequentially transform an input sequence
into increasingly abstract representations, with each layer's output serving as
the input to the subsequent layer. Let \(x^{(\ell)}\) denote the input to the
\(\ell^{th}\) layer, and \(h^{(\ell)} = f(x^{(\ell)})\) denote the output of the
\(\ell^{th}\) layer after applying the transformation function \(f(\cdot)\). The
output of the final layer, \(h^{(L)}\), is typically used for downstream tasks
(e.g., classification or natural language generation).

\begin{algorithm}[tb]
    \caption{Layer-Pruning Language Models}
    \label{alg:layer_pruning}
    \begin{algorithmic}[1]
        \STATE {\bfseries Input:} Model $M$ with $L$ layers, Number of layers to
        prune $n$, Dataset $D$ \STATE {\bfseries Output:} Pruned model $M'$
        \STATE Initialize $\ell^\star \gets \text{None}$, $d_{\min} \gets
        \infty$ \FOR{each layer $\ell$ from $1$ to $L-n$} \STATE $h^{(\ell)}(D)
        \gets \text{activation at } \ell \text{ with input } D$ \STATE
        $h^{(\ell+n)}(D) \gets \text{activation at } \ell+n \text{ with input }
        D$ \STATE Compute $d(h^{(\ell)}(D), h^{(\ell+n)}(D))$ using
        Eq.~\ref{eq:angular_cosine} \IF{$d(h^{(\ell)}(D), h^{(\ell+n)}(D)) <
        d_{\min}$} \STATE $d_{\min} \gets d(h^{(\ell)}(D), h^{(\ell+n)}(D))$
        \STATE $\ell^\star \gets \ell$ \ENDIF \ENDFOR \STATE Prune layers
        $\ell^\star$ to $\ell^\star + n - 1$ from $M$ \STATE Connect output of
        layer $\ell^\star$ to input of layer $\ell^\star + n$
        \STATE \textbf{return} pruned model $M'$
    \end{algorithmic}
\end{algorithm}



\begin{figure*}[t]
    \centering
    \includegraphics[width=\textwidth]{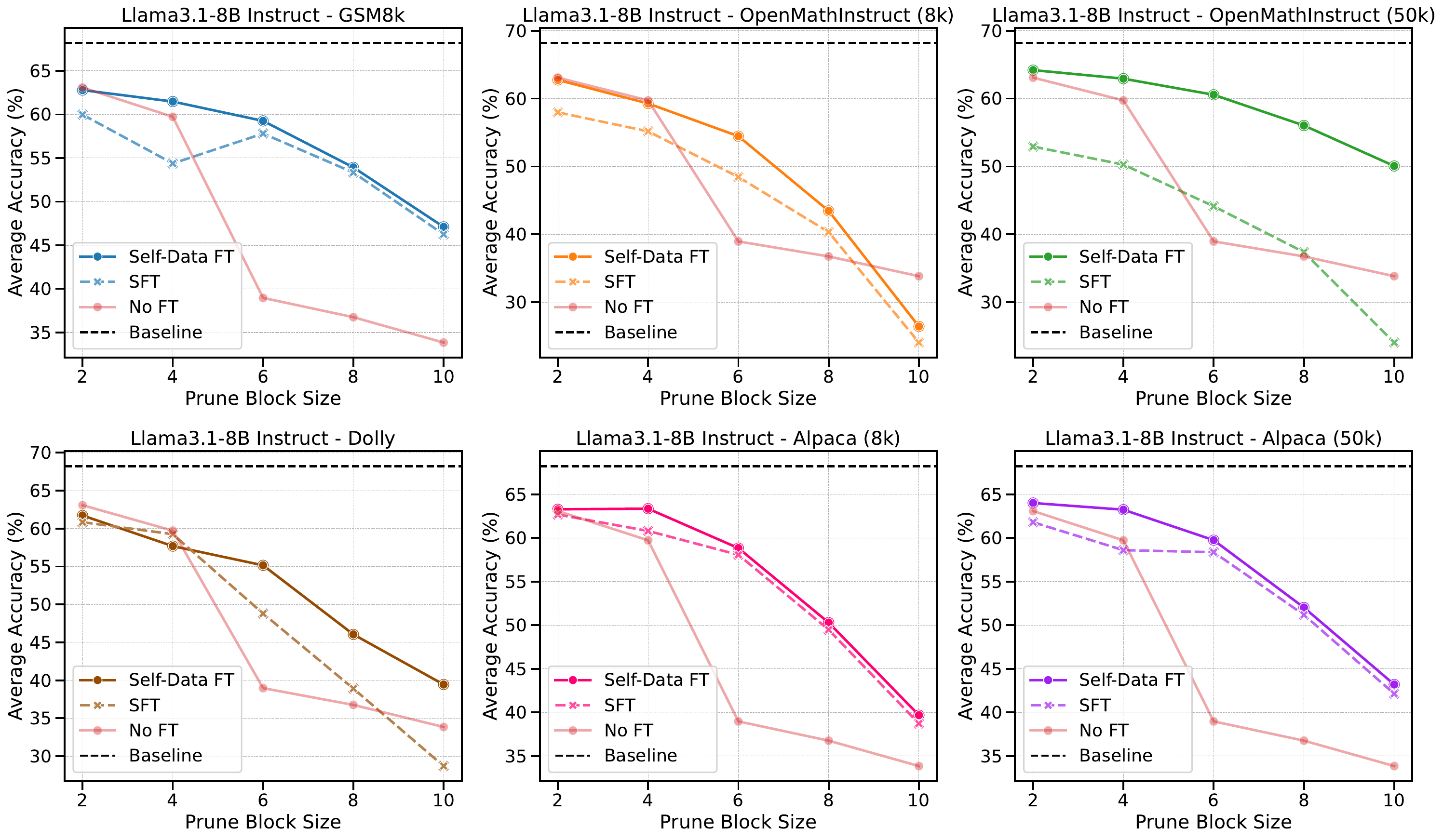}
    \vspace{-15pt}
    \caption{\textbf{Quality of pruned Llama3.1-8B Instruct models across various datasets and pruning block sizes.} 
    The plots show average accuracy across MMLU, GSM8k, ARC-C tasks for GSM8k,
    OpenMathInstruct, Dolly, and Alpaca under three strategies: Self-Data FT,
    SFT, and No FT. Self-Data FT consistently outperforms SFT and No FT, with
    the largest gains using OpenMathInstruct (50k).}
    \label{fig:average_performance}
    \vspace{-5pt}
\end{figure*}

\paragraph{Block Importance Metric} Recent literature has introduced various
metrics to evaluate the importance of layers within Transformer-based vision and
language models. For instance,
\citet{samragh2023weightsubcloningdirectinitialization} proposed a metric based
on the relative magnitude, $\left\lVert \frac{f(x^{(\ell)})}{x^{(\ell)} +
f(x^{(\ell)})} \right\rVert$ to measure the importance of a layer \(\ell\) by
characterizing its influence on the network's output. Here, \(x^{(\ell)}\)
represents the input to layer \(\ell\), and \(f(x^{(\ell)})\) denotes the
transformation applied by the layer. Additionally,
\citet{men2024shortgptlayerslargelanguage} introduced the Block Influence (BI)
score, which assumes that the degree to which a Transformer block alters hidden
states correlates with its importance. The BI score for the \(\ell^{th}\) block
is calculated as, $1 - \mathbb{E}_{X,i} \frac{x^{(\ell)}_i \cdot
x^{(\ell+1)}_i}{\left\lVert x^{(\ell)}_i \right\rVert_2 \left\lVert
x^{(\ell+1)}_i \right\rVert_2}$ where \(x^{(\ell)}_i\) represents the \(i^{th}\)
hidden state vector at layer \(\ell\), and \(x^{(\ell+1)}_i\) represents the
corresponding hidden state vector at the subsequent layer \(\ell+1\).
\citet{gromov2024unreasonableineffectivenessdeeperlayers} proposed using an
angular cosine metric to measure the similarity between layer outputs as a
criterion for pruning. This metric is based on the premise that layers producing
highly similar outputs can be pruned with minimal impact on the model's overall
quality. Both the BI and the angular cosine metric fundamentally use cosine
distance to assess the importance of layers or blocks of layers within a model.
However, based on our ablation studies in Section~\ref{sec:ablations}, we found
no significant difference in their effectiveness for layer pruning.
Consequently, we have opted to use the angular cosine metric
from~\citet{gromov2024unreasonableineffectivenessdeeperlayers} in our studies.
This metric allows us to effectively quantify and identify redundancy in the
model's layers. As described in Algorithm~\ref{alg:layer_pruning}, the pruning
process begins by selecting a block of consecutive layers, denoted by \(n\), for
potential removal. The choice of \(n\) directly influences the extent of
pruning, which has important implications for both the model's efficiency and
overall quality.
\vspace{-5pt}
\paragraph{Determine the Prune Block Size} To determine which layers to prune,
we calculate the angular distance between the activation outputs of layer
\(\ell\) and layer \(\ell+n\). For each potential starting layer \(\ell\), the
angular distance \(d(h^{(\ell)}(D), h^{(\ell+n)}(D))\) is computed using a
representative dataset \(D\), which may be a representative pre-training dataset
or one that is tailored to a specific downstream task. In our work, we use
RedPajama~\citep{together2023redpajama} as the representative dataset to
evaluate the sample distances (see more details in
Appendix~\ref{app:calib_structured_layer_pruning}). The angular distance metric,
$d(h^{(\ell)}(D), h^{(\ell+n)}(D))$, is formally defined as,
\begin{equation}
   \frac{1}{\pi} \arccos\left(\frac{h^{(\ell)}_T(D) \cdot h^{(\ell+n)}_T(D)}{\left\lVert h^{(\ell)}_T(D) \right\rVert \left\lVert h^{(\ell+n)}_T(D) \right\rVert}\right),
    \label{eq:angular_cosine}
\end{equation}
where \(h^{(\ell)}_T(D)\) and \(h^{(\ell+n)}_T(D)\) denote the activation
vectors at the final token position \(T\) of the input sequence, corresponding
to layers \(\ell\) and \(\ell+n\), respectively. The activations are normalized
using the \(L^2\)-norm \(\left\lVert \cdot \right\rVert\), which ensures a
consistent scale when comparing layer outputs. The choice to focus on \(T\) is
motivated by the autoregressive nature of Transformers, where the representation
of the final token encapsulates information from the entire input sequence due
to the causal attention mechanism.

\vspace{-5pt}
\paragraph{Identify Optimal Pruning Block} We identify the optimal block of
layers for pruning by minimizing the angular distance. Specifically, the
starting layer \(\ell^\star\) of the block is selected as follows,
\vspace{-5pt}
\[
\ell^\star(n) \equiv \argminA_\ell~ d(h^{(\ell)}(D), h^{(\ell+n)}(D)),
\]
where \(\ell^\star\) corresponds to the layer with the smallest angular distance
to its corresponding \(n\)-th successor layer. This optimization identifies a
block of layers that exhibit high redundancy, as measured by their similar
output activations. Pruning such a block is expected to have minimal impact on
the model's overall capacity. Once identified, layers from \(\ell^\star\) to
\(\ell^\star + n - 1\) are removed, and the model is restructured by directly
connecting the output of layer \(\ell^\star\) to the input of layer \(\ell^\star
+ n\). 

Aligning with~\citet{sreenivas2024llmpruningdistillationpractice}, we adopt a
one-shot pruning approach that balances implementation simplicity with
computational efficiency. Their method demonstrates that identifying and
removing redundant layers in one operation can achieve significant parameter
reduction while maintaining model quality when paired with appropriate
distillation. While our approach prioritizes this one-shot strategy, we
recognize that iterative pruning techniques, which progressively remove smaller
network portions with re-training between steps, may offer benefits in certain
architectural contexts. This represents a promising direction for future
research, particularly for models where more fine-grained control over the
accuracy-efficiency tradeoff is desired.


\vspace{-5pt}
\subsection{Self-Data Distillation for Pruned Models}

After pruning a Transformer, fine-tuning is typically required to adapt the
pruned model to specific downstream tasks. However, the fine-tuning process can
amplify catastrophic forgetting, especially when the fine-tuning data
distribution diverges from the original training distribution. To address this,
we utilize \textit{self-data distilled
fine-tuning}~\citep{Yang2024SelfDistillationBD}, which aligns the fine-tuning
dataset with the original model's learned distribution, to mitigate forgetting
and maintain model quality across a diverse range of tasks.

\vspace{-5pt}
\paragraph{Supervised Fine-tuning}
Given a pruned model \(M'\) with parameters \(\theta'\), supervised fine-tuning
aims to adapt the model to a specific downstream task \(t\) using a
task-specific dataset. For each example \((x^t, y^t)\) in the dataset, where
\(x^t\) is the input and \(y^t\) is the corresponding target output, the model
is fine-tuned by minimizing the negative log-likelihood of producing the correct
output \(y^t\) given the input \(x^t\) and the context \(c^t\) associated with
the task,
\vspace{-5pt}
\[
L_\text{SFT}(\theta') = -\log f_{\theta'}(y^t \mid c^t, x^t),
\]
where \(f_{\theta'}\) represents the pruned model with parameters \(\theta'\).
The objective is to align the model's output distribution with the distribution
of the task-specific data, thereby improving quality on the target task \(t\).
However, traditional supervised fine-tuning (SFT) can lead to catastrophic
forgetting~\citep{kotha2024understanding}, particularly in cases where the
task-specific data distribution diverges significantly from the original
training distribution.
\vspace{-5pt}
\paragraph{Self-Data Distilled Fine-tuning} 
First, the self-data distillation process begins with generating a distilled
dataset that aligns with the distribution of the original, unpruned model \(M\).
Specifically, for each example in the fine-tuning dataset, the original seed
model \(M\) is used to generate a distilled output \(\Tilde{y}\) by rewriting
the original response \(y^t\) as, $\Tilde{y} \sim f_{\theta}(y \mid c^t, x^t,
y^t)$, where \(f_{\theta}\) represents the original model with parameters
\(\theta\). This distilled output \(\Tilde{y}\) is designed to stay within the
distribution of the original model, thereby minimizing the risk of catastrophic
forgetting. Following~\citet{Yang2024SelfDistillationBD}, to ensure the quality
of the distilled output, a conditional selection process is applied, 
\vspace{-5pt}
\[
\Tilde{y}' = \begin{cases} \Tilde{y} & \text{if } \text{Extract}(\Tilde{y}) =
y^t, \\
    y^t & \text{otherwise}. \end{cases}
\]
This ensures that the distilled responses retain essential characteristics, such
as correctness in structured tasks (e.g., mathematical reasoning). Once the
distilled dataset is prepared, the pruned model \(M'\) with parameters
\(\theta'\) undergoes supervised fine-tuning. The fine-tuning process is defined
by the following objective,
\vspace{-5pt}
\[
L_\text{Self-Data FT}(\theta') = -\log f_{\theta'}(\Tilde{y}' \mid c^t, x^t),
\]
where \(f_{\theta'}\) represents the pruned model fine-tuned on the distilled
dataset. This objective helps align the pruned model with the distilled data
distribution, thereby reducing the impact of catastrophic forgetting compared to
standard supervised fine-tuning. Self-data distillation is key to minimizing
quality loss after pruning, significantly improving retention of model
capabilities. While it may not fully preserve the original model's quality, it
offers an effective balance between efficiency and accuracy in LLMs, as
evidenced by the results presented in the ablation studies in
Section~\ref{sec:ablations}.

%% file: tex_files/ablation.tex
\begin{figure}[t]
  \centering
  \includegraphics[width=0.45\columnwidth, trim=0mm 0mm 26mm 0mm, clip]{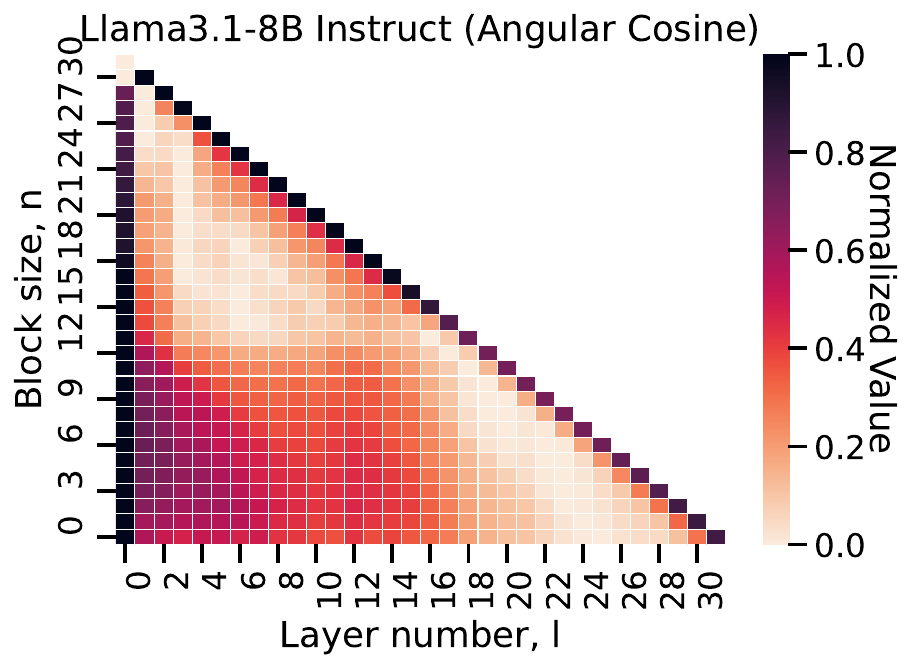}
  \includegraphics[width=0.49\columnwidth, trim=15mm 0mm 0mm 0mm, clip]{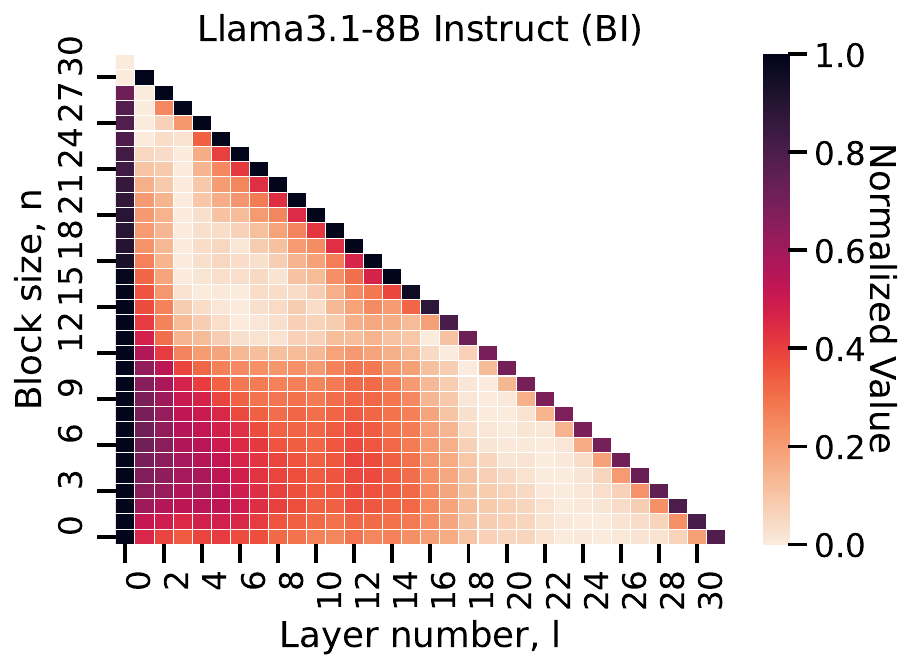}
  \vspace{-5pt}
  \caption{Comparison of Llama3.1-8B Instruct using (left) \textit{angular
  cosine} and (right) \textit{block influence} (BI) score metrics. Both metrics
  highlight inherent redundancy in the middle layers of Llama3.1-8B Instruct,
  suggesting that these layers can be pruned with minimal impact on overall
  model quality.}
  \label{fig:metric_comparison}
  \vspace{-5pt}
\end{figure}

\begin{figure*}[t]
  \centering
  \hfill
  \includegraphics[width=0.31\textwidth]{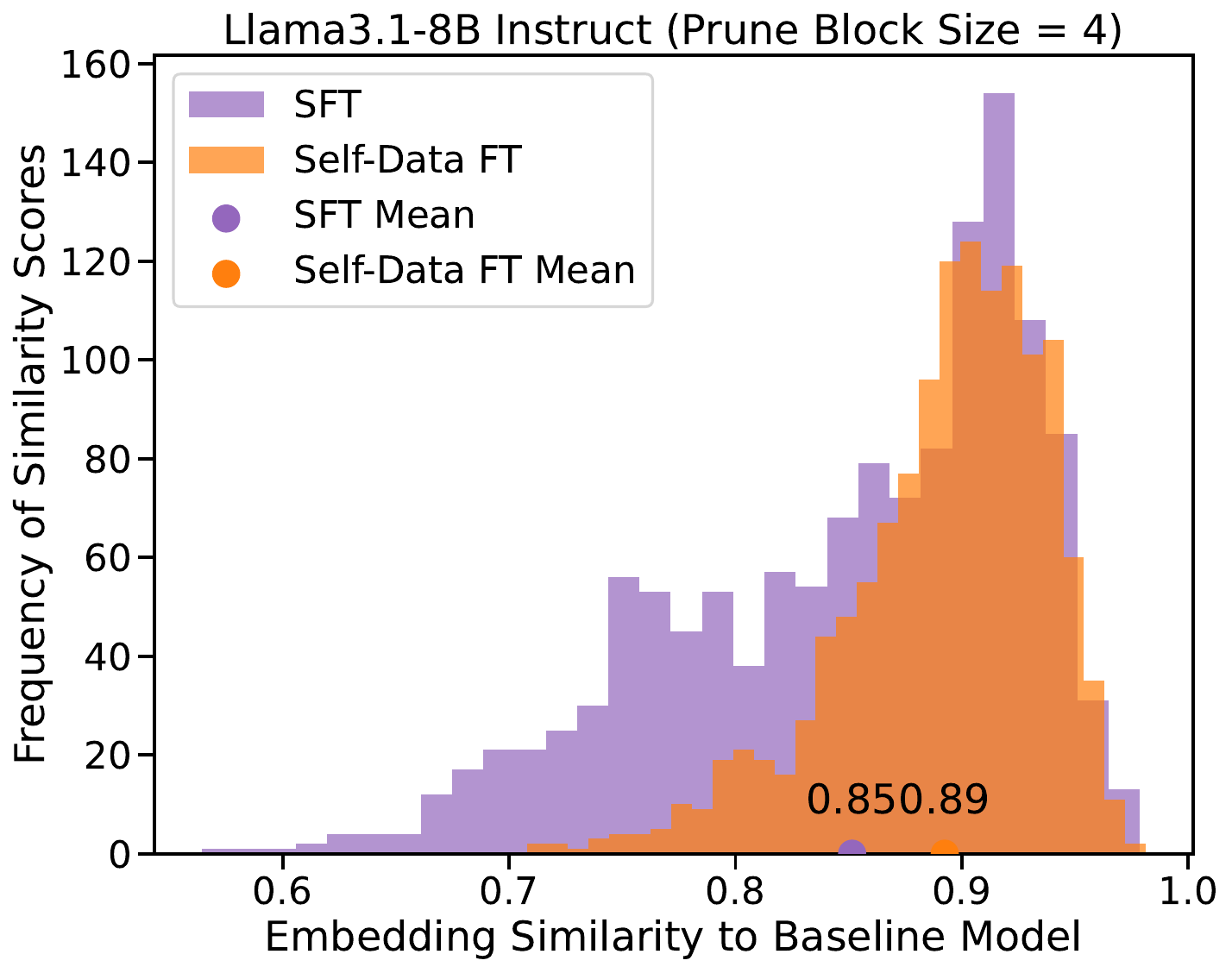}
  \hfill
  \includegraphics[width=0.31\textwidth]{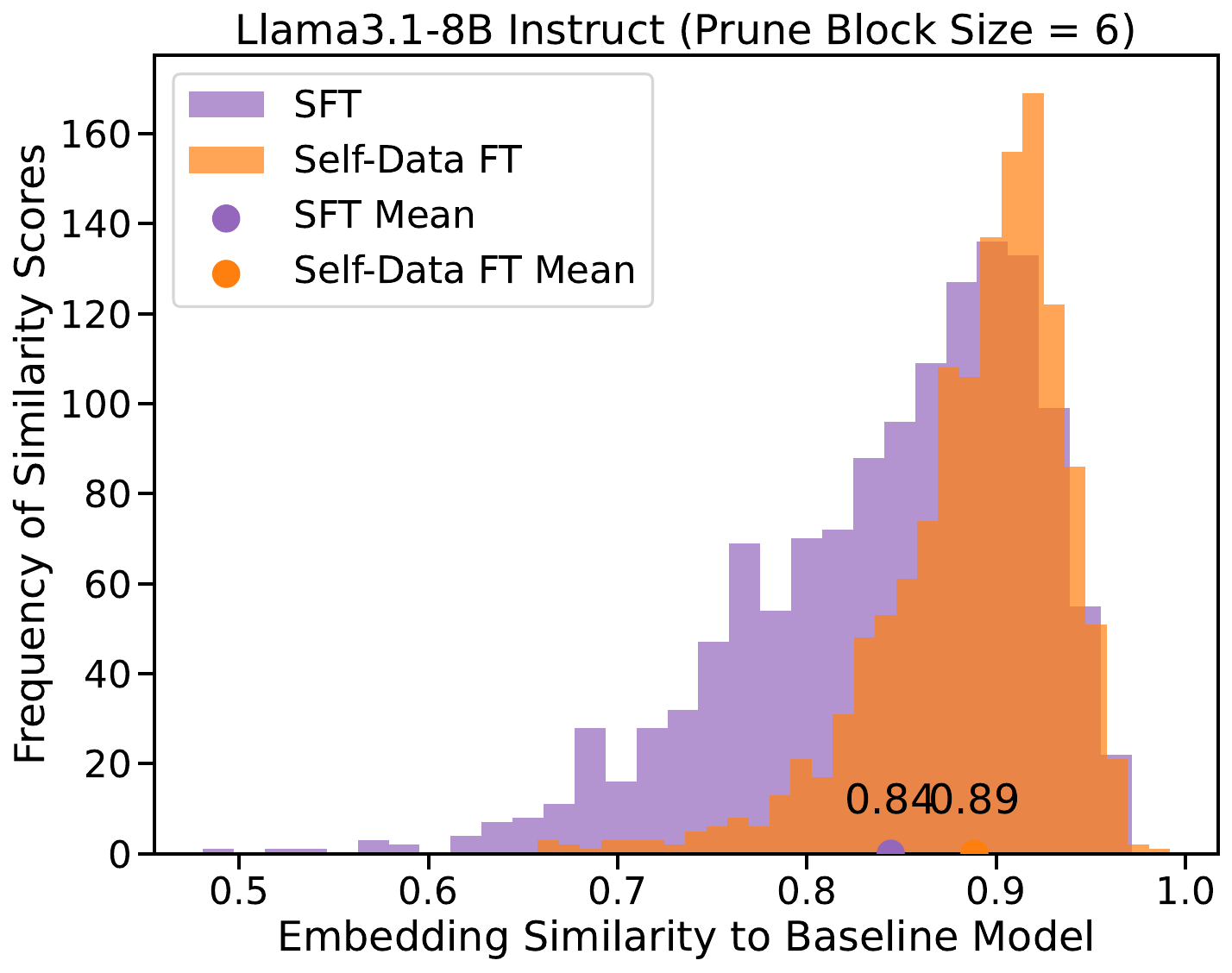}
  \hfill
  \includegraphics[width=0.31\textwidth]{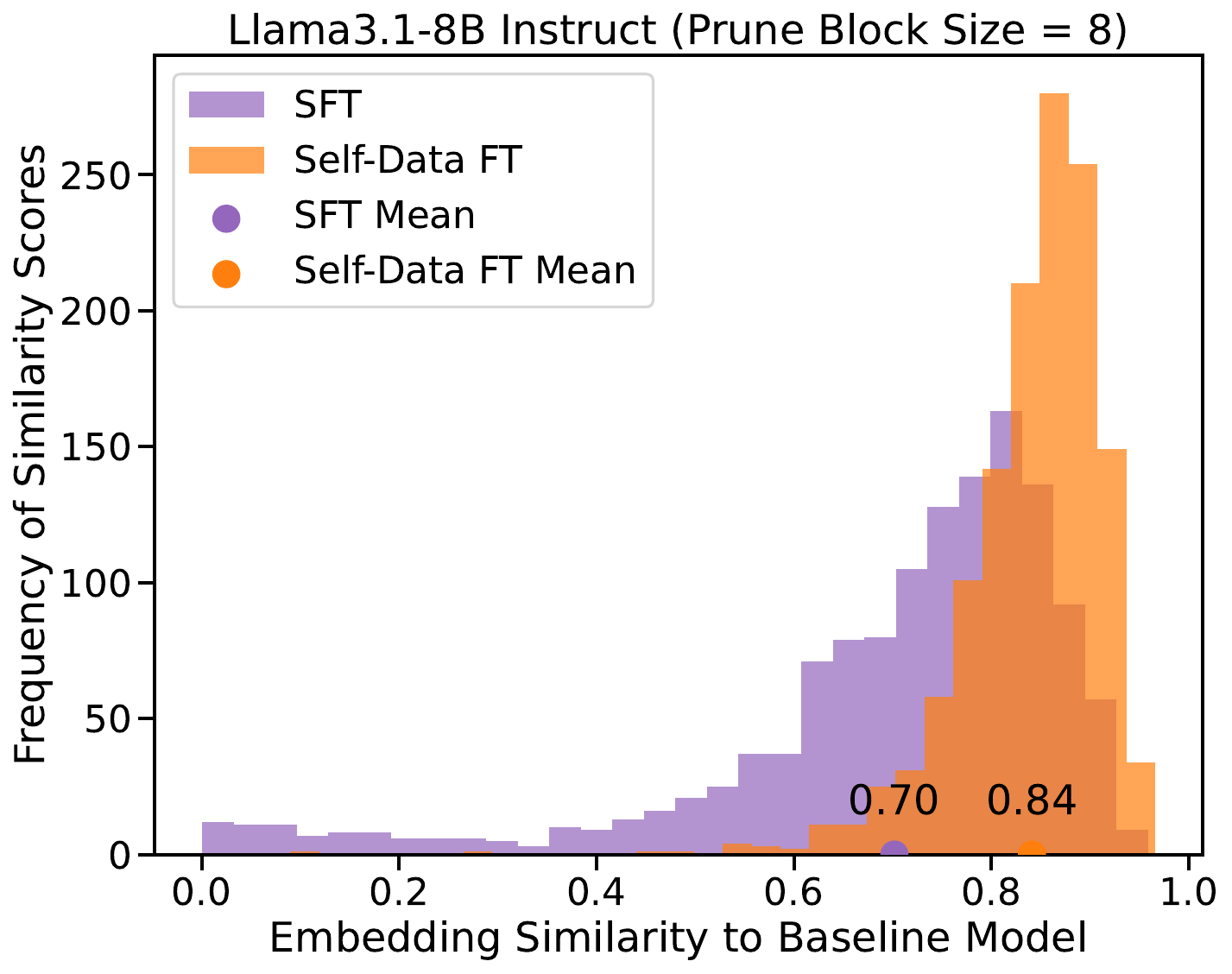}
  \hfill
  \vspace{-5pt}
  \caption{\textbf{Distribution of embedding similarities on GSM8k test
  dataset.} We present the distribution of embedding similarities after
  fine-tuning a structurally pruned variant of the Llama3.1-8B Instruct model on
  50k samples of OpenMathInstruct. We compute the cosine similarity between the
  sentence embeddings of the pruned models and those generated by the original
  Llama3.1-8B Instruct model. The plots show: (left) 28 decoder layers (prune
  block size = 4), (center) 26 decoder layers (prune block size = 6), and
  (right) 24 decoder layers (prune block size = 8). Self-Data Distilled
  Fine-Tuning (Self-Data FT) achieves higher similarity to the original baseline
  model, indicating a reduced distribution shift compared to Supervised
  Fine-Tuning (SFT). }
  \label{fig:metric_comparison}
  \vspace{-5pt}
\end{figure*}

\vspace{-5pt}
\section{Ablation Studies}
\label{sec:ablations}
In this section, we examine key factors affecting the quality of pruned
Llama3.1-8B Instruct~\citep{dubey2024llama3herdmodels} models. Our experiments
assess the impact of layer importance metrics, pruning block sizes, and
fine-tuning strategies. We compare BI and angular cosine metrics for determining
layer redundancy and analyze how these choices influence pruning outcomes. In
addition, we evaluate the effectiveness of self-data distilled fine-tuning
across various datasets, showing its ability to recover model quality
post-pruning, outperforming standard supervised fine-tuning methods.
\vspace{-5pt}
\subsection{Effect of Layer Importance Metric}
We investigated two layer importance metrics, BI and angular cosine distance, to
guide pruning decisions in the Llama3.1-8B Instruct model (see
Figure~\ref{fig:metric_comparison}). Both metrics assess the redundancy of
layers by measuring the cosine distance between their inputs and outputs but
differ in their focus. BI quantifies the change in hidden states across layers,
while angular cosine distance emphasizes output similarity between consecutive
layers. 
Despite minor differences in the middle layers, both metrics produced comparable
pruning results across block sizes. We ultimately chose the angular cosine
metric for its computational efficiency, as it uses cosine similarity on only
the last token, whereas BI operates over the entire sequence, adding
computational overhead. This efficiency makes it more scalable for LLMs, and its
direct measure of output similarity aligns with our intuition that layers
producing highly similar outputs are likely redundant, making it a practical
tool for structured pruning in this study.


\begin{table*}
  \caption{\textbf{Model quality results for pruned Llama3.1-8B Instruct models
  across various pruning block sizes and fine-tuning strategies.} Average
  accuracy is reported on the OpenLLM Leaderboard v1, along with the recovery
  percentage relative to the baseline model. Self-data distillation consistently
  outperforms outperforms standard supervised fine-tuning (SFT).}
  
  \label{tab:llama3_pruning_results}
  \centering
  \begin{adjustbox}{max width=\textwidth}
  \begin{tabular}{cccccccccccccccccc}
      \toprule
      \begin{tabular}[c]{@{}c@{}}Prune\\ Block Size\end{tabular}&
      \begin{tabular}[c]{@{}c@{}}Model\\ Savings\end{tabular} &
      \begin{tabular}[c]{@{}c@{}}Fine-tuning\\ Method\end{tabular} & Dataset &
      \begin{tabular}[c]{@{}c@{}}ARC-C\\ (25-shot)\end{tabular} &
      \begin{tabular}[c]{@{}c@{}}HellaSwag\\ (10-shot)\end{tabular}  &
      \begin{tabular}[c]{@{}c@{}}TruthfulQA\\ (0-shot)\end{tabular}  &
      \begin{tabular}[c]{@{}c@{}}MMLU\\ (5-shot)\end{tabular}  &
      \begin{tabular}[c]{@{}c@{}}Winogrande\\ (5-shot)\end{tabular}  &
      \begin{tabular}[c]{@{}c@{}}GSM8k\\ (5-shot)\end{tabular}  &
      \begin{tabular}[c]{@{}c@{}}Avg.\\ Score\end{tabular}  &
      \begin{tabular}[c]{@{}c@{}}Avg.\\ Recovery \end{tabular} \\
      \midrule
      Baseline &  -    & \multicolumn{1}{l}{No FT}  &   & 60.92 & 80.16 & 54.02
      & 68.15 & 77.58 & 75.59 & 69.40 & - \\
      \midrule
      \multirow{3}{*}{2} & \multirow{3}{*}{
        \begin{tabular}[c]{@{}c@{}}5.23\%\\ (7.61B) \end{tabular}} &
        \multicolumn{1}{l}{No FT} & & \textbf{58.71}	& \textbf{78.12}	& \textbf{53.21}	& 63.15	& \textbf{76.72} &	67.39	& \textbf{66.22}	& \textbf{95.41\%} \\
      &  & \multicolumn{1}{l}{SFT}  & \multicolumn{1}{l}{OpenMathInstruct}  
      & 52.91	& 73.24	& 51.77	& 60.91	& 73.75	& 44.95 &	59.59	& 85.85\% \\
      & & \multicolumn{1}{l}{Self-Data Distillation}  &
      \multicolumn{1}{l}{OpenMathInstruct}  & 56.41 & 75.03	& 50.23 & 	\textbf{66.17}	& 75.18	& \textbf{69.97}	& 65.51	& 94.37\% \\
      \midrule
      \multirow{3}{*}{4} & \multirow{3}{*}{
        \begin{tabular}[c]{@{}c@{}}10.86\%\\ (7.16B) \end{tabular}} &
        \multicolumn{1}{l}{No FT} & & 55.20 & 75.70 & \textbf{52.40} &
        \textbf{67.79} & \textbf{75.29} & 56.18 & 64.06 & 92.31\% \\
      &  & \multicolumn{1}{l}{SFT}  & \multicolumn{1}{l}{OpenMathInstruct}  &
      51.62 & 71.70 & 49.40 & 61.65 & 73.25 & 44.35 & 58.66 & 84.52\% \\
      & & \multicolumn{1}{l}{Self-Data Distillation}  &
      \multicolumn{1}{l}{OpenMathInstruct}  & \textbf{53.93} & 74.27 & 50.61 &
      65.34 & 74.90 & \textbf{69.44} & \textbf{64.75} & \textbf{93.29\%} \\
      \midrule
      \multirow{3}{*}{6} & \multirow{3}{*}{
        \begin{tabular}[c]{@{}c@{}}16.30\%\\ (6.72B) \end{tabular}} &
        \multicolumn{1}{l}{No FT} & & 49.49 & 68.72 & \textbf{53.63} &
        \textbf{67.42} & 70.40 & 1.29 & 51.82 & 74.67\% \\
      & & \multicolumn{1}{l}{SFT}  & \multicolumn{1}{l}{OpenMathInstruct}  &
      46.93 & 68.51 & 50.81 & 59.98 & 70.01 & 43.82 & 56.68 & 81.66\% \\
      &  & \multicolumn{1}{l}{Self-Data Distillation}  &
      \multicolumn{1}{l}{OpenMathInstruct}  & \textbf{50.02} & \textbf{71.57} &
      53.14 & 64.96 & \textbf{73.64} & \textbf{66.64} & \textbf{63.33} &
      \textbf{91.24\%} \\
      \midrule
      \multirow{3}{*}{8} & \multirow{3}{*}{
        \begin{tabular}[c]{@{}c@{}}21.73\%\\ (6.29B) \end{tabular}} &
        \multicolumn{1}{l}{No FT} & & 44.71 & 61.22 & \textbf{56.11} &
        \textbf{65.57} & 65.98 & 0.00 & 48.93 & 70.50\% \\
      &  & \multicolumn{1}{l}{SFT}  & \multicolumn{1}{l}{OpenMathInstruct}  &
      42.15 & 64.49 & 54.69 & 60.65 & 66.38 & 29.64 & 53.00 & 76.37\% \\
      &  & \multicolumn{1}{l}{Self-Data Distillation}  &
      \multicolumn{1}{l}{OpenMathInstruct}  & \textbf{46.67} & \textbf{65.70} &
      53.32 & 64.87 & \textbf{71.27} & \textbf{57.70} & \textbf{59.92} &
      \textbf{86.38\%} \\
      \midrule
      \multirow{3}{*}{10} & \multirow{3}{*}{
        \begin{tabular}[c]{@{}c@{}}27.16\%\\ (5.85B) \end{tabular}}  &
        \multicolumn{1}{l}{No FT} & & 37.46 & 54.45 & \textbf{55.06} & 64.09 &
        67.25 & 0.00 & 46.39 & 66.83\% \\
      &  & \multicolumn{1}{l}{SFT}  & \multicolumn{1}{l}{OpenMathInstruct}  &
      39.33 & 57.38 & 51.47 & 57.44 & 64.48 & 15.39 & 47.58 & 68.56\% \\
      &  & \multicolumn{1}{l}{Self-Data Distillation}  &
      \multicolumn{1}{l}{OpenMathInstruct}  & \textbf{40.88} & \textbf{61.11} &
      53.46 & \textbf{64.54} & \textbf{70.88} & \textbf{44.58} & \textbf{55.91}
      & \textbf{80.56\%} \\
      \bottomrule
  \end{tabular}
  \end{adjustbox}
  \vspace{-5pt}
\end{table*}

\begin{table*}[t]
  \caption{\textbf{Model quality results for pruned Mistral-7B-v0.3 Instruct
  models across various pruning block sizes and fine-tuning strategies.} Average
  accuracy is reported on the OpenLLM Leaderboard v1, along with the recovery
  percentage relative to the baseline model. Self-data distillation consistently
  outperforms standard supervised fine-tuning (SFT).}
  
  \label{tab:mistral_pruning_results}
  \centering
  \begin{adjustbox}{max width=\textwidth}
  \begin{tabular}{cccccccccccccccccc}
      \toprule
      \begin{tabular}[c]{@{}c@{}}Prune\\ Block Size\end{tabular}&
      \begin{tabular}[c]{@{}c@{}}Model\\ Savings\end{tabular} &
      \begin{tabular}[c]{@{}c@{}}Fine-tuning\\ Method\end{tabular} & Dataset &
      \begin{tabular}[c]{@{}c@{}}ARC-C\\ (25-shot)\end{tabular} &
      \begin{tabular}[c]{@{}c@{}}HellaSwag\\ (10-shot)\end{tabular}  &
      \begin{tabular}[c]{@{}c@{}}TruthfulQA\\ (0-shot)\end{tabular}  &
      \begin{tabular}[c]{@{}c@{}}MMLU\\ (5-shot)\end{tabular}  &
      \begin{tabular}[c]{@{}c@{}}Winogrande\\ (5-shot)\end{tabular}  &
      \begin{tabular}[c]{@{}c@{}}GSM8k\\ (5-shot)\end{tabular}  &
      \begin{tabular}[c]{@{}c@{}}Avg.\\ Score\end{tabular}  &
      \begin{tabular}[c]{@{}c@{}}Avg.\\ Recovery \end{tabular} \\
      \midrule 
      Baseline &  -    & \multicolumn{1}{l}{No FT}  &   & 63.39 & 84.52 & 59.65
      & 61.99 & 78.37 & 48.75 & 66.11 & - \\
      \midrule 
      \multirow{3}{*}{2} & \multirow{3}{*}{
        \begin{tabular}[c]{@{}c@{}}6.02\%\\ (6.83B) \end{tabular}} &
        \multicolumn{1}{l}{No FT} & & 60.24	& 62.31	& 58.57	& 62.11 & \textbf{78.77}	&
        32.75	& 59.12	& 89.43\% \\
      &  & \multicolumn{1}{l}{SFT}  & \multicolumn{1}{l}{OpenMathInstruct}  &
      54.01	& \textbf{79.38}	& \textbf{53.71}	& 57.61	& 74.82	& 34.04	& 58.93	& 89.13\% \\
      & & \multicolumn{1}{l}{Self-Data Distillation}  &
      \multicolumn{1}{l}{OpenMathInstruct}  & \textbf{53.93} & 74.27 & 50.61 &
      \textbf{65.34} & 74.90 & \textbf{69.44} & \textbf{64.75} & \textbf{93.29\%} \\
      \midrule
      \multirow{3}{*}{4} & \multirow{3}{*}{
        \begin{tabular}[c]{@{}c@{}}12.03\%\\ (6.39B) \end{tabular}} &
        \multicolumn{1}{l}{No FT} & & \textbf{57.17} & \textbf{78.89} &
        \textbf{57.44} & \textbf{61.36} & \textbf{77.19} & 9.33 & 56.90 &
        86.06\% \\
      &  & \multicolumn{1}{l}{SFT}  & \multicolumn{1}{l}{OpenMathInstruct}  &
      52.65	& 75.29	& 53.16	& 58.21	& 74.51	& 27.22	& 56.84	& 85.97\% \\
      & & \multicolumn{1}{l}{Self-Data Distillation}  &
      \multicolumn{1}{l}{OpenMathInstruct}  & 55.29	& 75.77	& 53.33	& 60.12	&
      75.22	& \textbf{34.65}	& \textbf{59.06}	& \textbf{89.34\%} \\
      \midrule
      \multirow{3}{*}{6} & \multirow{3}{*}{
        \begin{tabular}[c]{@{}c@{}}18.06\%\\ (5.96B) \end{tabular}} &
        \multicolumn{1}{l}{No FT} & & \textbf{50.25} & \textbf{72.67} &
        \textbf{55.85} & \textbf{61.56} & \textbf{75.45} & 0.83 & 52.77 &
        79.82\% \\
      & & \multicolumn{1}{l}{SFT}  & \multicolumn{1}{l}{OpenMathInstruct}  &
      47.52	& 70.15	& 51.86 &	58.29 &	74.91 & 20.47	& 53.87	& 81.47\% \\
      &  & \multicolumn{1}{l}{Self-Data Distillation}  &
      \multicolumn{1}{l}{OpenMathInstruct}  & 48.98	& 70.67	& 52.28	& 59.56	&
      72.22	& \textbf{28.13}	& \textbf{55.31}	& \textbf{83.65\%} \\
      \midrule
      \multirow{3}{*}{8} & \multirow{3}{*}{
        \begin{tabular}[c]{@{}c@{}}24.08\%\\ (5.52B) \end{tabular}} &
        \multicolumn{1}{l}{No FT} & & 41.04 & 65.27 & \textbf{58.10} &
        \textbf{61.05} & 71.74 & 0.00 & 49.53 & 74.92\% \\
      &  & \multicolumn{1}{l}{SFT}  & \multicolumn{1}{l}{OpenMathInstruct}  &
      43.09	& 66.11	& 52.74	& 58.52 &	\textbf{72.21}	& 9.48	& 50.35	& 76.16\%
      \\
      &  & \multicolumn{1}{l}{Self-Data Distillation}  &
      \multicolumn{1}{l}{OpenMathInstruct}  & \textbf{45.14}	& \textbf{67.09} &
      53.61 &	59.81 &	71.43	& \textbf{19.03}	& \textbf{52.68}	&
      \textbf{79.69\%} \\
      \midrule 
      \multirow{3}{*}{10} & \multirow{3}{*}{
        \begin{tabular}[c]{@{}c@{}}30.10\%\\ (5.08B) \end{tabular}}  &
        \multicolumn{1}{l}{No FT} & & 34.30 & 52.14 & \textbf{58.67} & 35.76 &
        65.11 & 0.00 & 41.00 & 62.01\% \\
      &  & \multicolumn{1}{l}{SFT}  & \multicolumn{1}{l}{OpenMathInstruct}  &
      39.33	& 59.07	& 52.07	& 57.49	& 68.75	& 6.07& 47.13	& 71.29\% \\
      &  & \multicolumn{1}{l}{Self-Data Distillation}  &
      \multicolumn{1}{l}{OpenMathInstruct}  & \textbf{39.76} &	\textbf{59.25} &
      51.61	& \textbf{58.42} &	\textbf{69.61}	& \textbf{14.39} &
      \textbf{48.84}	& \textbf{73.87\%} \\
      \bottomrule
  \end{tabular}
  \end{adjustbox}
  \vspace{-5pt}
\end{table*}

\vspace{-5pt}
\subsection{Analysis on Self-Data Distilled Datasets}
We assess the role of fine-tuning datasets in the self-data distillation process
for recovering quality in pruned Llama3.1-8B Instruct models, comparing
LoRA~\citep{hu2022lora} fine-tuning on standard versus self-distilled datasets.
We fine-tuned the pruned models on a range of open-source datasets, including
GSM8k (math word problems), Dolly~\citep{DatabricksBlog2023DollyV2} (open-domain
conversation), OpenMathInstruct~\citep{toshniwal2024openmath} (math and
reasoning), and Alpaca~\citep{alpaca} (instruction-following), with a primary
focus on reasoning-heavy tasks. Therefore, we evaluated the orginal baseline and
pruned models' accuracy on ARC-C (25-shot), GSM8k (5-shot), and
MMLU~\citep{hendrycks2021measuring} (5-shot) tasks using the
LM-eval-harness~\citep{eval-harness}. We provide additional details on the
experimental setup in Appendix~\ref{app:experiment_details}, and extended
results from our fine-tuning ablation studies can be found in
Appendix~\ref{app:ablations_extended}.

The results, presented in Figure~\ref{fig:average_performance}, show that
self-data distillation offers significant quality improvements over SFT and
one-shot pruned models, particularly with larger datasets like the 50k-sample
OpenMathInstruct. Self-data distilled models achieve 5-10\% higher accuracy than
models without any fine-tuning for pruning block sizes above four. Across all
block sizes, they also outperform models fine-tuned with SFT, demonstrating the
effectiveness of self-data distillation in recovering a substantial portion of
the model's quality post-pruning. Our ablations revealed a strong correlation
between dataset size and quality recovery, with larger datasets consistently
outperforming smaller ones. The improvements being most pronounced with medium
to large block sizes, where self-data distillation notably enhanced
generalization, especially in reasoning tasks. These findings highlight the
importance of dataset scale and the self-data distillation process, with the
50k-sample OpenMathInstruct dataset delivering the best overall results, and
will be the focus of subsequent experiments. 

\vspace{-5pt}
\subsection{Mitigating Catastrophic Forgetting}
Figure~\ref{fig:metric_comparison} illustrates the advantages of self-data
distilled fine-tuning (Self-Data FT) over SFT in mitigating catastrophic
forgetting after pruning. We study the similarities between Llama3.1-8B Instruct
models fine-tuned on a supervised 50k-sample OpenMathInstruct dataset and a
self-data distilled version of the same dataset. The plots compares the sentence
embeddings of the two models' generated responses on the GSM8k task to the
baseline unpruned Llama3.1-8B Instruct model at various prune block sizes.
Self-Data FT maintains a narrower distribution of embedding similarities, with
higher mean scores, indicating better preservation of the original model's
learned representations. In contrast, SFT yields a wider spread and reduced mean
in similarity scores, indicative of a heavier-tailed distribution. This shift
suggests a more pronounced distributio shift, thereby increasing the risk of
quality degradation. The tighter distribution in Self-Data FT highlights its
ability to retain model quality across tasks, while the wider distribution in
SFT suggests a greater distribution shift, leading to higher risk of
catastrophic forgetting (see Appendix~\ref{app:catastrophic_forgetting}). This
highlights Self-Data FT as a more effective method for mitigating model quality
degradation post-pruning.

%% file: tex_files/empirical_results.tex
\begin{table}[ht]
    \caption{\textbf{Model quality results for pruned Llama3.1-8B Instruct
    models across various pruning block sizes.} Average accuracy is reported on
    the OpenLLM Leaderboard v1, along with the recovery percentage relative to
    the baseline model. Self-data distillation with model merging (MM)
    consistently outperforms self-data distillation with mixed datasets.}
    
    \label{tab:llama3_pruning_results_mm}
    \centering
    \begin{adjustbox}{max width=\columnwidth}
    \begin{tabular}{cccccccccccccccccc}
        \toprule
        \begin{tabular}[c]{@{}c@{}}Prune\\ Block Size\end{tabular}&
        \begin{tabular}[c]{@{}c@{}}Fine-tuning\\ Method\end{tabular} & Dataset &
        \begin{tabular}[c]{@{}c@{}}Avg.\\ Score\end{tabular}  &
        \begin{tabular}[c]{@{}c@{}}Avg.\\ Recovery \end{tabular} \\
        \midrule
        Baseline & \multicolumn{1}{l}{No FT}  &   & 69.40 & - \\
        \midrule
        \multirow{3}{*}{4} & \multicolumn{1}{l}{No FT} & & 64.06 & 92.31\% \\
        &  \multicolumn{1}{l}{Self-Data Distillation}  &
        \multicolumn{1}{l}{OpenMathInstruct + Alpaca}  & 65.27 & 94.04\% \\
        & \multicolumn{1}{l}{Self-Data Distillation + MM}  &
        \multicolumn{1}{l}{OpenMathInstruct + Alpaca}  & \textbf{65.84} &
        \textbf{94.86\%} \\
        \midrule
        \multirow{3}{*}{6} & \multicolumn{1}{l}{No FT} & & 51.82 & 74.67\% \\
        &  \multicolumn{1}{l}{Self-Data Distillation}  &
        \multicolumn{1}{l}{OpenMathInstruct + Alpaca}  & 63.77 & 91.88\% \\
        &   \multicolumn{1}{l}{Self-Data Distillation + MM}  &
        \multicolumn{1}{l}{OpenMathInstruct + Alpaca}  &  \textbf{64.75} &
        \textbf{93.30\%} \\
        \midrule
        \multirow{3}{*}{8} & \multicolumn{1}{l}{No FT} & & 48.93 & 70.50\% \\
        &   \multicolumn{1}{l}{Self-Data Distillation}  &
        \multicolumn{1}{l}{OpenMathInstruct + Alpaca}  & 59.37 & 85.54\% \\
        &   \multicolumn{1}{l}{Self-Data Distillation + MM}  &
        \multicolumn{1}{l}{OpenMathInstruct + Alpaca}  & \textbf{61.24} &
        \textbf{88.24\%} \\
        \midrule
        \multirow{3}{*}{10} & \multicolumn{1}{l}{No FT} & & 46.39 & 66.83\% \\
        &   \multicolumn{1}{l}{Self-Data Distillation}  &
        \multicolumn{1}{l}{OpenMathInstruct + Alpaca}  & 55.85 & 80.47\% \\
        &  \multicolumn{1}{l}{Self-Data Distillation + MM}  &
        \multicolumn{1}{l}{OpenMathInstruct + Alpaca}  & \textbf{56.08} &
        \textbf{80.70\%} \\
        \bottomrule
    \end{tabular}
    \end{adjustbox}
    \vspace{-5pt}
  \end{table}

  \begin{table}[t]
    \caption{\textbf{Model quality results for pruned Mistral-7B-v0.3 Instruct
    models across various pruning block sizes.} Average accuracy is reported on
    the OpenLLM Leaderboard v1, along with the recovery percentage relative to
    the baseline model. Self-data distillation with model merging (MM)
    consistently outperforms self-data distillation with mixed datasets.}
    
    \label{tab:mistral_pruning_results_mm}
    \centering
    \begin{adjustbox}{max width=\columnwidth}
    \begin{tabular}{cccccccccccccccccc}
        \toprule
        \begin{tabular}[c]{@{}c@{}}Prune\\ Block Size\end{tabular}&
        \begin{tabular}[c]{@{}c@{}}Fine-tuning\\ Method\end{tabular} & Dataset &
        \begin{tabular}[c]{@{}c@{}}Avg.\\ Score\end{tabular}  &
        \begin{tabular}[c]{@{}c@{}}Avg.\\ Recovery \end{tabular} \\
        \midrule
        Baseline & \multicolumn{1}{l}{No FT}  &   & 66.11 & - \\
        \midrule
        \multirow{3}{*}{4} & \multicolumn{1}{l}{No FT} & & 56.90 & 86.06\% \\
        &  \multicolumn{1}{l}{Self-Data Distillation}  &
        \multicolumn{1}{l}{OpenMathInstruct + Alpaca}  & 59.04 & 89.30\% \\
        & \multicolumn{1}{l}{Self-Data Distillation + MM}  &
        \multicolumn{1}{l}{OpenMathInstruct + Alpaca}  & \textbf{59.57} &
        \textbf{90.10\%} \\
        \midrule
        \multirow{3}{*}{6} & \multicolumn{1}{l}{No FT} & & 52.77 &	79.82\% \\
        &  \multicolumn{1}{l}{Self-Data Distillation}  &
        \multicolumn{1}{l}{OpenMathInstruct + Alpaca}  & 56.26 & 85.09\% \\
        &   \multicolumn{1}{l}{Self-Data Distillation + MM}  &
        \multicolumn{1}{l}{OpenMathInstruct + Alpaca}  & \textbf{56.41 } &
        \textbf{85.32\%} \\
        \midrule
        \multirow{3}{*}{8} & \multicolumn{1}{l}{No FT} & & 49.53 & 74.92\% \\
        &   \multicolumn{1}{l}{Self-Data Distillation}  &
        \multicolumn{1}{l}{OpenMathInstruct + Alpaca}  & 52.78 &	79.83\% \\
        &   \multicolumn{1}{l}{Self-Data Distillation + MM}  &
        \multicolumn{1}{l}{OpenMathInstruct + Alpaca}  & \textbf{54.14} &
        \textbf{81.89\%} \\
        \midrule
        \multirow{3}{*}{10} & \multicolumn{1}{l}{No FT} & & 41.00 & 62.01\% \\
        &   \multicolumn{1}{l}{Self-Data Distillation}  &
        \multicolumn{1}{l}{OpenMathInstruct + Alpaca}  & 48.86 & 73.90\% \\
        &  \multicolumn{1}{l}{Self-Data Distillation + MM}  &
        \multicolumn{1}{l}{OpenMathInstruct + Alpaca}  &  \textbf{50.04} &
        \textbf{75.69\%} \\
        \bottomrule
    \end{tabular}
    \end{adjustbox}
    \vspace{-5pt}
  \end{table}

\vspace{-5pt}
\section{Empirical Results}\label{sec:empirical_results}
We evaluated the quality of Llama3.1-8B Instruct and Mistral-7B-v0.3 Instruct
models pruned at various block sizes under three fine-tuning strategies: no
fine-tuning (No FT), supervised fine-tuning (SFT), and our proposed self-data
distillation. As shown in Table~\ref{tab:llama3_pruning_results}, pruned models
without fine-tuning experience substantial accuracy losses, particularly at
larger block sizes (e.g., a 46.39\% average score at block size 10),
highlighting the critical need for post-pruning adaptation. SFT improves
quality, with an average recovery of 81.66\% at block size 6, but struggles on
reasoning-heavy tasks like GSM8k and ARC-C. In contrast, self-data distillation
significantly enhances quality recovery, achieving 91.24\% at block size 6, with
GSM8k accuracy reaching 66.64\% (compared to 43.82\% with SFT). Even at block
size 10, self-data distillation maintains an 80.56\% recovery, outperforming
SFT's 68.56\%. Similarly, in Table~\ref{tab:mistral_pruning_results}, the
Mistral-7B models exhibit improved quality recovery across all pruning block
sizes.  These findings establish self-data distillation as an effective method
for preserving model quality post-pruning, particularly for reasoning-intensive
tasks, making it essential for large-scale compression.

\begin{table*}[ht]
    \caption{\textbf{Evaluation of speculative decoding performance using self-data distilled Llama3.1-8B Instruct draft models for the Llama3.1-70B Instruct target model.} 
    The draft models generate 6 speculative tokens plus 1 base token from the
    target model, with accepted lengths measured across all tasks from
    Spec-Bench. Results are shown for various pruning block sizes, where
    self-data distillation with MM consistently improves over the no fine-tuning
    (No FT) baseline, with higher average accepted lengths indicating better
    speculative capabilities and better alignment with the target model. At a
    prune block size of 10, we achieve 27.16\% FLOP savings, and the average
    accepted length across tasks (final column) improves by 1.70.}
    
    \label{tab:llama_sd}
    \centering
    \begin{adjustbox}{max width=\textwidth}
    \begin{tabular}{cccccccccccccccccc}
        \toprule
        \multicolumn{4}{c}{} & \multicolumn{7}{c}{\textbf{Average Accepted
        Length on Spec-Bench}  } \\
        \cmidrule(lr){5-11}
        \multicolumn{1}{c}{\begin{tabular}[c]{@{}c@{}}Target\\
        Model\end{tabular}}&
        \multicolumn{1}{c}{\begin{tabular}[c]{@{}c@{}}Prune\\ Block
        Size\end{tabular}} &
        \multicolumn{1}{c}{\begin{tabular}[c]{@{}c@{}}Model\\
        Savings\end{tabular}} &
        \multicolumn{1}{c}{\begin{tabular}[c]{@{}c@{}}Fine-tuning\\
        Method\end{tabular}} & \multicolumn{1}{c}{MT-Bench} &
        \multicolumn{1}{c}{Translation} & \multicolumn{1}{c}{Summarization} &
        \multicolumn{1}{c}{QA} & \multicolumn{1}{c}{Math Reasoning} &
        \multicolumn{1}{c}{RAG} & \multicolumn{1}{c}{Average} \\
        \midrule 
        \multirow{11}{*}{\begin{tabular}[c]{@{}c@{}}Llama3.1-70B Instruct \\
        (70B) \end{tabular}} & Baseline &  -    & \multicolumn{1}{l}{No FT} &
        5.03	& 4.25	& 4.72 &	4.86&	5.86&	4.73&	5.01 \\
        \cmidrule{2-11}
        & \multirow{2}{*}{4} & \multirow{2}{*}{
          \begin{tabular}[c]{@{}c@{}}10.86\%\\ (7.16B) \end{tabular}} &
          \multicolumn{1}{l}{No FT} &  3.50 &	3.18 &	3.62 &	3.16 &	4.26 &
          3.56 &	3.55 \\
        & & & \multicolumn{1}{l}{Self-Data Distillation + MM}  
         & 4.26	&3.71	&4.13	&3.53	&5.56	&4.13		&
         \textbf{4.22}$_{\textcolor{ForestGreen}{\uparrow
         \textbf{\text{0.67}}}}$  \\
        \cmidrule{2-11}
        & \multirow{2}{*}{6} & \multirow{2}{*}{
          \begin{tabular}[c]{@{}c@{}}16.30\%\\ (6.72B) \end{tabular}} &
          \multicolumn{1}{l}{No FT} & 2.77 & 2.40 &	2.72 & 2.35 & 3.13 & 2.65 &
          2.67 \\
        & & & \multicolumn{1}{l}{Self-Data Distillation + MM}  
         & 3.93	& 3.40	& 3.91	& 3.14	& 5.30	& 3.83	&
          \textbf{3.92}$_{\textcolor{ForestGreen}{\uparrow
          \textbf{\text{1.25}}}}$  \\
          \cmidrule{2-11}
        & \multirow{2}{*}{8} & \multirow{2}{*}{
          \begin{tabular}[c]{@{}c@{}}21.73\%\\ (6.29B) \end{tabular}} &
          \multicolumn{1}{l}{No FT} & 2.38	& 2.26	& 2.70 & 2.11 & 2.89 &  2.55
          & 2.40  \\
        & & & \multicolumn{1}{l}{Self-Data Distillation + MM}  
         & 	3.63	&3.17&	3.77	&2.91&	4.99 &	3.41 &
         \textbf{3.65}$_{\textcolor{ForestGreen}{\uparrow
         \textbf{\text{1.25}}}}$   \\
         \cmidrule{2-11}
        & \multirow{2}{*}{10} & \multirow{2}{*}{
          \begin{tabular}[c]{@{}c@{}}27.16\%\\ (5.85B) \end{tabular}}  &
          \multicolumn{1}{l}{No FT}  & 1.58	& 1.48 &	1.50	& 1.48 & 1.63 & 1.49
          & 1.55 \\
        & & & \multicolumn{1}{l}{Self-Data Distillation + MM}  
         & 3.17	& 2.76 & 3.21	& 2.49 & 4.59 &	3.30 &
         \textbf{3.25}$_{\textcolor{ForestGreen}{\uparrow
         \textbf{\text{1.70}}}}$ \\
        \bottomrule
    \end{tabular}
    \end{adjustbox}
    \vspace{-5pt}
  \end{table*}

\begin{table*}
    \caption{\textbf{Evaluation of speculative decoding performance using self-data distilled Mistral-7B-v0.3 Instruct draft models for the Mistral Large 2 (123B) target model.} 
    Results on Spec-Bench are shown for various pruning block sizes,
    highlighting the effectiveness of self-data distillation paired with model
    merging (MM). Self-data distillation with MM consistently improves over the
    no fine-tuning baseline, with higher average accepted lengths indicating
    better speculative capabilities and better alignment with the target model.
    At a prune block size of 10, we achieve 30.10\% FLOP savings, and the
    average accepted length across tasks (final column) improves by 1.62.}
    
    \label{tab:mistral_sd}
    \centering
    \begin{adjustbox}{max width=\textwidth}
    \begin{tabular}{cccccccccccccccccc}
        \toprule
        \multicolumn{4}{c}{} & \multicolumn{7}{c}{\textbf{Average Accepted
        Length on Spec-Bench} } \\
        \cmidrule(lr){5-11}
        \multicolumn{1}{c}{\begin{tabular}[c]{@{}c@{}}Target\\
        Model\end{tabular}}&
        \multicolumn{1}{c}{\begin{tabular}[c]{@{}c@{}}Prune\\ Block
        Size\end{tabular}} &
        \multicolumn{1}{c}{\begin{tabular}[c]{@{}c@{}}Model\\
        Savings\end{tabular}} &
        \multicolumn{1}{c}{\begin{tabular}[c]{@{}c@{}}Fine-tuning\\
        Method\end{tabular}} & \multicolumn{1}{c}{MT-Bench} &
        \multicolumn{1}{c}{Translation} & \multicolumn{1}{c}{Summarization} &
        \multicolumn{1}{c}{QA} & \multicolumn{1}{c}{Math Reasoning} &
        \multicolumn{1}{c}{RAG} & \multicolumn{1}{c}{Average} \\
        \midrule 
        \multirow{11}{*}{\begin{tabular}[c]{@{}c@{}}Mistral Large 2\\ (123B)
        \end{tabular}} & Baseline &  -    & \multicolumn{1}{l}{No FT}     & 3.98
        &4.36	&3.70&	3.50	&4.85	&3.85		&4.04 \\
        \cmidrule{2-11}
        & \multirow{2}{*}{4} & \multirow{2}{*}{
          \begin{tabular}[c]{@{}c@{}}12.03\%\\ (6.83B) \end{tabular}} &
          \multicolumn{1}{l}{No FT} &  3.40 &	3.76 &	3.28 &	2.77 &	4.28 &
          3.22 &	3.45 \\
        & & & \multicolumn{1}{l}{Self-Data Distillation + MM}  
         & 3.70	& 4.06	& 3.53	& 2.99	& 4.81	& 3.44		&
         \textbf{3.75}$_{\textcolor{ForestGreen}{\uparrow
         \textbf{\text{0.30}}}}$  \\
        \cmidrule{2-11}
        & \multirow{2}{*}{6} & \multirow{2}{*}{
          \begin{tabular}[c]{@{}c@{}}18.06\%\\ (5.96B) \end{tabular}} &
          \multicolumn{1}{l}{No FT} &  2.97 & 3.30 &	2.93	&2.36	&3.78 &2.82	&
          3.03 \\
        & & & \multicolumn{1}{l}{Self-Data Distillation + MM}  
          & 3.53	& 3.86	& 3.41	& 2.79	& 4.72	& 3.30	&
          \textbf{3.60}$_{\textcolor{ForestGreen}{\uparrow
          \textbf{\text{0.58}}}}$  \\
          \cmidrule{2-11}
        & \multirow{2}{*}{8} & \multirow{2}{*}{
          \begin{tabular}[c]{@{}c@{}}24.08\%\\ (5.52B) \end{tabular}} &
          \multicolumn{1}{l}{No FT} & 1.98	& 2.16	& 2.03	& 1.76	& 2.66	&
          1.98	& 2.10  \\
        & & & \multicolumn{1}{l}{Self-Data Distillation + MM}  
         & 3.37	& 3.68	& 3.25	& 2.58	& 4.62	& 3.10	&
         \textbf{3.43}$_{\textcolor{ForestGreen}{\uparrow
         \textbf{\text{1.34}}}}$   \\
         \cmidrule{2-11}
        & \multirow{2}{*}{10} & \multirow{2}{*}{
          \begin{tabular}[c]{@{}c@{}}30.10\%\\ (5.06B) \end{tabular}}  &
          \multicolumn{1}{l}{No FT}  &  1.50	& 1.55 & 1.54& 1.45	& 1.73	& 1.59
          & 1.56 \\
        & & & \multicolumn{1}{l}{Self-Data Distillation + MM}  
         & 3.15 &	3.25	& 2.97	& 2.38	& 4.48	& 2.87	&
         \textbf{3.18}$_{\textcolor{ForestGreen}{\uparrow
         \textbf{\text{1.62}}}}$ \\
        \bottomrule
    \end{tabular}
    \end{adjustbox}
    \vspace{-5pt}
  \end{table*}

\vspace{-5pt}
\subsection{Self-Data Distillation with Model Merging}

We extend self-data distillation by introducing \textit{model merging} using
Spherical Linear Interpolation (SLERP)~\citep{slerpKen1985}. The model merging
technique aims to synthesize complementary knowledge from models fine-tuned on
distinct datasets, potentially leading to improved generalization and robustness
after pruning. While various model merging techniques, such as
TIES~\citep{yadav2023tiesmerging}, DARE-TIES~\citep{yu2024language}, and Linear
Merge~\citep{nagarajan2019}, have been proposed, prior empirical studies have
demonstrated SLERP to be particularly
effective~\citep{aakanksha2024mixdatamergemodels}, as it effectively balances
the dual objectives of maintaining alignment and preserving overall model
quality. SLERP enables smooth interpolation between two model parameters along
the shortest path on a high-dimensional sphere, preserving the geometric
properties of the parameter space (see Appendix~\ref{app:model_merging} for
details).

Building on the gains achieved by self-data distillation in recovering quality
post-pruning, we investigate whether merging models fine-tuned on diverse
datasets can provide further improvements. We perform model merging within each
model family by merging Llama3.1-8B Instruct models fine-tuned on
OpenMathInstruct and Alpaca, and similarly merging Mistral-7B models fine-tuned
on the same datasets, as these configurations delivered the best results in our
ablations in Section~\ref{sec:ablations}.  To establish a robust baseline for
comparison against model merging, we combined the OpenMathInstruct and Alpaca
datasets, fine-tuning the pruned models on this interleaved dataset. The total
number of training samples was kept consistent across both approaches to ensure
a fair evaluation. 

As presented in Table~\ref{tab:llama3_pruning_results_mm} for Llama3.1 models
and Table~\ref{tab:mistral_pruning_results_mm} for Mistral-7B models, our
results indicate that self-data distillation combined with model merging via
SLERP achieves the highest recovery of model quality across all pruning block
sizes. Specifically, at a pruning block size of 6, the merged Llama3.1 model
demonstrates a 93.30\% recovery of its original quality, surpassing the 91.88\%
recovery observed with models trained on a mixed dataset. Similarly, the
Mistral-7B model achieves an 85.32\% recovery at block size 6 through merging,
slightly exceeding the 85.09\% recovery obtained with the mixed dataset
approach. These results highlight the consistency and effectiveness of self-data
distillation with model merging in retaining model quality across varying
pruning configurations. These findings suggest that SLERP-based model merging
not only mitigates pruning-related quality loss but also improves
generalization, particularly on tasks such as GSM8k and ARC-C. 

Based on these empirical results, we derive several important insights for
effective model pruning and quality recovery. First, using robust,
instruction-tuned base models is critical as they provide a stable foundation
for pruning and subsequent fine-tuning operations. We observe that both
Llama3.1-8B Instruct and Mistral-7B-v0.3 Instruct models benefit significantly
from our approach precisely because they start from a well-calibrated
instruction-following state. Second, incorporating diverse, representative
datasets during fine-tuning and distillation—as demonstrated by our
OpenMathInstruct and Alpaca combination—minimizes distribution mismatches and
improves generalization across varied tasks. This diversity is particularly
important for maintaining performance on reasoning-intensive benchmarks after
aggressive pruning. Finally, our method highlights the flexibility enabled by
decoupling pre-training and fine-tuning phases, allowing self-data distillation
to effectively transfer knowledge across various contexts while maintaining
alignment with the original model's capabilities. This architectural separation
ensures that pruned models can be effectively adapted to specific deployment
scenarios without compromising their fundamental abilities.

%% file: tex_files/speculative_decode.tex
\vspace{-5pt}
\subsection{Speculative Decoding and Self-Data Distillation}

Speculative decoding is founded on two key insights into the inference dynamics
of LLMs. First, generating tokens with a smaller draft
model reduces computational cost by offloading initial token prediction from the
larger target model. Second, LLM inference is primarily constrained by memory
bandwidth, with latency bottlenecks arising from parameter memory access rather
than arithmetic operations~\citep{pattersonBandwidth2004,
shazeer2019fasttransformerdecodingwritehead}. By adapting principles from
speculative execution, speculative decoding aims to prioritize the verification
of pre-drafted tokens, thereby minimizing frequent memory operations and
enhancing overall inference efficiency. Despite its potential, speculative
decoding raises several challenges that require deeper exploration. Central
among these is the need to design an effective drafting mechanism that optimally
balances speculative accuracy with drafting
efficiency~\citep{xia-etal-2023-speculative,
li2024eaglespeculativesamplingrequires}. The effectiveness of this approach
depends on two critical factors: the accuracy of the draft model, measured by
the average number of accepted tokens per decoding step, and the drafting
latency itself~\citep{stern2018, xia-etal-2023-speculative}. Achieving a
trade-off between high speculative accuracy and low latency remains a
significant challenge, as both elements are essential for maximizing the overall
speedup.

While not the main focus of this work, speculative decoding is a natural
application of our method. In this context, integrating self-data distillation
with speculative decoding presents a viable solution to improve inference
performance. Self-data distillation aligns the draft model more closely with the
target model by generating a distilled dataset that preserves semantic
consistency even after pruning. This integration not only enhances the draft
model's speculative capabilities but also mitigates quality degradation,
allowing for more efficient and accurate token validation during inference.
Consequently, the combination of speculative decoding and self-data distillation
emerges as a promising approach for improving both inference speed and model
quality retention across diverse tasks.

The draft-and-verify setup in speculative decoding involves two phases:
\textit{drafting} and \textit{verification}. In the drafting phase, a smaller
draft model \(M_p\) speculates multiple future tokens. Given an input sequence
\(\{x_1,\ldots,x_n\}\), it generates \(K\) speculative tokens \(\tilde{x}_j\) by
sampling from probability distributions \(p_j\), where \(\{p_1, \ldots, p_K\} =
\text{Draft}(x_{\leq n}, M_p)\) and \(\tilde{x}_j \sim p_j\) for \(j =
\{1,\ldots,K\}\). In the verification phase, the larger target model \(M_q\)
verifies these tokens by computing probabilities for the drafted tokens as \(q_j
= M_q(x \mid x_{\leq n}, \tilde{x}_{<j})\) for \(j = \{1, \ldots, K+1\}\). Each
token \(\tilde{x}_j\) is then checked against a criterion
\(\text{Verify}(\tilde{x}_j, p_j, q_j)\), where in our experimental setup,
verification is performed using \(\tilde{x}_j = \arg \max q_j\). If a token
fails verification, the first failing token \(\tilde{x}_c\) is corrected via
\(\text{Correct}(p_c, q_c)\), specifically as \(x_{n+c} \leftarrow \arg \max
q_c\), and subsequent tokens are discarded. If all tokens pass verification, an
additional token \(x_{n+K+1}\) is sampled from the target model. This process
repeats until the [EOS] token is generated or a maximum sequence length is
reached.


\vspace{-5pt}
\paragraph{Experimental Setup} Our speculative decoding experiments use pruned,
self-data distilled models with different pruning block sizes $n \in \{4, 6, 8,
10 \}$. These models are evaluated on the Spec-Bench
dataset~\citep{xia-etal-2024-unlocking}, a comprehensive benchmark that covers
categories including multi-turn chat, translation, summarization, question
answering (QA), mathematical reasoning, and retrieval-augmented generation
(RAG). There are a total of 480 samples in Spec-Bench, where each category in
Spec-Bench consists of 80 randomly selected instances drawn from six
well-established datasets: MT-Bench~\citep{lianminMTBench2024}, WMT14
DE-EN~\citep{nallapati-etal-2016-abstractive}, CNN/Daily
Mail~\citep{nallapati-etal-2016-abstractive}, Natural
Questions~\citep{kwiatkowski-etal-2019-natural},
GSM8K~\citep{cobbe2021trainingverifierssolvemath}, and
DPR~\citep{karpukhin-etal-2020-dense}.

We report the overall average accepted token lengths to measure the
effectiveness of speculative decoding. Our speculative decoding configuration
generates \(K = 6\) speculative tokens \(\{\tilde{x}_1, \ldots, \tilde{x}_6\}\)
using the draft model \(M_p\), while the target model \(M_q\) concurrently
verifies these tokens and generates an additional token \(x_{n+1}\). We explore
two setups, 1$)$ pruned self-data distilled Llama3.1 models speculating for the
Llama3.1-70B Instruct~\citep{dubey2024llama3herdmodels} target model, and 2$)$ pruned
self-data distilled Mistral-7B Instruct models speculating for the Mistral
Large 2 (123B)~\citep{mistrallarge2} target model.

\vspace{-5pt}
\paragraph{Results} Our evaluations indicate that self-data distillation
enhances speculative capabilities by improving alignment with the target model.
Notably, with larger prune block sizes (e.g., 10), self-data distillation paired
with model merging achieves higher average accepted lengths across all
categories, demonstrating better generalization and more efficient speculative
decoding. For instance, in Table~\ref{tab:llama_sd} pruned Llama3.1 models at
block size 10 (i.e., 5.85B resulting in 27.16\% FLOPS savings) show an average
accepted length improvement of 1.70 tokens. Similarly, in
Table~\ref{tab:mistral_sd}, we observe Mistral models at block size 10 (i.e.,
5.06B resulting in 30.10\% FLOPs savings) achieve a gain of 1.62 tokens. While
the improvements are evident across general tasks, substantial gains are
particularly observed in mathematical reasoning. Pruned Llama3.1 and Mistral-7B
models at block size 10 demonstrate an average accepted length increase of 2.96
and 2.75 tokens, respectively, on mathematical reasoning benchmarks. These
results show the effectiveness of self-data distillation in preserving complex
reasoning capabilities, even under aggressive pruning. Such findings highlight
the interplay between speculative decoding and self-data distillation, where the
combination optimizes inference efficiency and model quality retention across a
broad range of tasks, with notable robustness in challenging reasoning domains.

%% file: tex_files/related_work.tex
\vspace{-5pt}
\section{Related Work}

\paragraph{Pruning for Model Compression} Pruning is a well-established method
for reducing the complexity of overparameterized models in both computer vision
and NLP~\citep{lecunBrain1989, babak1993obs}. It is typically classified into
\textit{structured} and \textit{unstructured} pruning. Unstructured pruning
removes individual weights and can achieve high compression rates in LLMs,
particularly when paired with hardware accelerators like the Cerebras
CS-3~\citep{liehotchips, pmlr-v235-thangarasa24a} or Neural Magic
DeepSparse~\citep{deepsparse_neuralmagic_2021}, which exploit sparsity for
significant speedups. However, without specialized infrastructure, unstructured
pruning can result in inefficient acceleration. Structured pruning, which
removes entire channels, layers, or attention heads, is more effective in models
with architectural redundancy, but can degrade model quality, especially in
complex tasks which require multi-step reasoning~\citep{kurtic2023ziplm,
ma2023llmpruner, sun2024transformerlayerspainters}.

To address these challenges, several metrics have been developed to guide
pruning decisions more effectively. For instance, Shortened
Llama~\citep{kim2024shortenedllamadepthpruning} demonstrated that depth pruning
(removing layers) can be as effective as width pruning (removing units within
layers), or even a combination of both. The Block Influence (BI)
score~\citep{men2024shortgptlayerslargelanguage}, applied in
Llama-2~\citep{touvron2023llama2openfoundation}, measures block importance by
evaluating changes in hidden state magnitudes. Additionally, the angular cosine
similarity metric~\citep{gromov2024unreasonableineffectivenessdeeperlayers}
identifies layers with redundant activations, allowing for selective pruning in
models such as Llama-2 and Mistral~\citep{jiang2023mistral7b}.
\citet{gromov2024unreasonableineffectivenessdeeperlayers} also proposed a
healing method using low-rank adapters~\citep{hu2022lora} to recover lost
quality. Despite these advancements, pruning LLMs still results in sharp
accuracy degradation~\citep{sun2024transformerlayerspainters}, and traditional
recovery methods such as fine-tuning or re-pretraining are
resource-intensive~\citep{xia2024sheared,
sreenivas2024llmpruningdistillationpractice}. Our self-data distillation
approach extends this by leveraging the unpruned model to generate a distilled
dataset for fine-tuning the pruned model, improving semantic alignment and
mitigating the quality degradation caused by pruning. While combining this with
standard KD techniques could further improve generalization, we leave that for
future work.

\vspace{-5pt}
\paragraph{Distillation}
Knowledge distillation~\citep{hinton2015distilling} is a widely-used model
compression technique where a smaller student model learns from a larger teacher
model, enabling efficient model quality retention. In NLP, KD has been applied
in various contexts to align student models with teacher
outputs~\citep{liang2021mixkd, gu2024minillm, agarwal2024onpolicy}, hidden
states~\citep{jiao-etal-2020-tinybert}, and attention
mechanisms~\citep{wang-etal-2021-minilmv2}. Recent work on
Llama3.2~\citep{metaLlama32} extends this by using logits from larger Llama3.1
models (e.g., 8B, 70B) as token-level targets during pre-training, allowing the
smaller models (e.g., 1B, 3B) to achieve superior quality compared to training
from scratch~\citep{metaLlama32}. Supervised fine-tuning (SFT) has been widely
employed in various self-distillation frameworks to train student models using
sequences generated by teacher LLMs~\citep{sun2023principledriven,
wang-etal-2023-self-instruct, zelikman2022star}.
\citet{Yang2024SelfDistillationBD} investigated self-distillation as a way to
alleviate distribution shifts, improving model quality during SFT while
improving generalization across tasks. Our self-data distillation method builds
on these techniques by leveraging the original unpruned model to generate a
distilled dataset for fine-tuning the pruned model. This enhances semantic
alignment and mitigates the quality degradation seen after pruning. Furthermore,
while our approach can be combined with KD methods to enhance generalization and
recover quality while lowering computational costs, we leave the exploration of
such combinations for future work.

\vspace{-5pt}
\paragraph{Catastrophic Forgetting}
One of the major challenges of pruning and distillation techniques in LLMs is
catastrophic forgetting, where a model loses its previously learned capabilities
during fine-tuning~\citep{kotha2024understanding, pmlr-v162-korbak22a}.
Regularization techniques such as Elastic Weight
Consolidation~\citep{KirkpatrickEWC2017} aim to alleviate this by controlling
parameter updates, but are task-dependent and require careful
tuning~\citep{huang-etal-2021-continual}. Architecture-based methods, which
allocate separate parameters for each task~\citep{razdaibiedina2023progressive},
preserve task-specific knowledge but add complexity and overhead, reducing the
overall efficiency of model compression. Replay-based
techniques~\citep{Ostapenko2022ContinualLW, NEURIPS2019_fa7cdfad,
Sun2019LAMOLLM} store data subsets from previous tasks for rehearsal, either
through direct storage or synthesis via generative models. However, these
methods demand substantial memory to store large datasets and are often
impractical due to privacy concerns or lack of access to past data. Our
self-data distilled fine-tuning approach avoids these challenges by aligning the
fine-tuning dataset with the original model's learned distribution, preserving
knowledge across tasks without requiring new parameters or architectural
changes. This method offers a robust solution for mitigating catastrophic
forgetting while maintaining model quality after pruning.

%% file: tex_files/conclusion.tex
\vspace{-5pt}
\section{Conclusion}

In conclusion, we introduce \textit{self-data distilled fine-tuning} as an
effective method to mitigate quality degradation in pruned LLMs, addressing
catastrophic forgetting while preserving alignment with the model's original
data distribution. Our approach consistently outperforms standard supervised
fine-tuning, demonstrating superior accuracy recovery post-pruning across
various downstream tasks on the OpenLLM Leaderboard v1. Additionally, model
merging via SLERP further enhances recovery, achieving significant quality
retention. We also show that our method scales with dataset size, where larger
self-distilled datasets lead to improved quality recovery. Moreover, integrating
self-data distillation with speculative decoding not only enhances token
acceptance rates but also reduces inference latency, offering an effective
strategy for deploying pruned LLMs efficiently. These findings highlight
self-data distilled fine-tuning as a critical tool for maintaining high model
quality post-pruning, offering an efficient solution for model compression.
Future work may involve integrating self-data distilled fine-tuning with
complementary model compression techniques such as sparsity, quantization or
teacher distillation, potentially yielding greater efficiency without
sacrificing model quality. Extending these methodologies to next-generation LLM
architectures presents a promising avenue for unlocking additional computational
efficiency and model robustness.

%% file: tex_files/appendix.tex
\section{Experimental Setup Details}
\label{app:experiment_details}

\subsection{Baseline Model} In our ablation studies in
Section~\ref{sec:ablations}, we used Llama3.1-8B
Instruct\footnote{\url{https://huggingface.co/meta-llama/Llama-3.1-8B-Instruct}}\citep{dubey2024llama3herdmodels}
as the baseline model for all experiments. This model comprises a total of 32
decoder layers, pretrained on a diverse array of instruction-following datasets.
This model was chosen for its strong generalization performance across a wide
range of natural language processing (NLP) tasks, making it an ideal candidate
for studying the impact of structured pruning and fine-tuning. The 8B model size
strikes a balance between computational efficiency and model quality, providing
a robust foundation for the experiments in this study. Hence, served as the
starting point for our structured layer pruning ablations and experiments in
Sections~\ref{sec:ablations} and~\ref{sec:empirical_results}, respectively. 

In Section~\ref{sec:empirical_results}, to further understand the efficacy of
our methodology on other LLMs, we also applied it to Mistral-7B Instruct
v0.3~\footnote{\url{https://huggingface.co/mistralai/Mistral-7B-Instruct-v0.3}},
an instruct fine-tuned version of Mistral-7B-v0.3. This most recent version of
Mistral-7B~\citep{jiang2023mistral7b}, compared to Mistral-7B-v0.2, includes an
extended vocabulary of 32,768 tokens, supports a v3 tokenizer, and enables
function calling.

\subsection{Structured Layer Pruning }
In this study, we focus on structured layer pruning of decoder layers to reduce
the computational footprint of the LLM while maintaining its quality.
Specifically, we prune in block sizes of \{2, 4, 6, 8, 10\} layers,
corresponding to \{30, 28, 26, 24, 22\} decoder layers, respectively. Each block
size reduction effectively removes a group of layers from the original
architecture, creating progressively smaller models. These pruned models allow
us to systematically evaluate the trade-offs between computational efficiency
(fewer layers) and the accuracy on various downstream tasks. By examining
multiple block sizes, we analyze how varying degrees of pruning impact model
quality, especially in the context of \textit{self-data distilled fine-tuning},
our proposed methodology.

\subsection{Calibration Dataset for Structured Layer Pruning}
\label{app:calib_structured_layer_pruning}
In structured pruning, selecting a suitable calibration dataset is critical for
effectively identifying and removing redundant layers without sacrificing model
quality. This study examines the impact of different calibration datasets: C4,
RedPajama, and SlimPajama on computing the angular cosine distance block
importance metric for Llama3.1-8B Instruct, as shown in
Table~\ref{tab:calibration_ablation_grouped_blocksize}. We found that all three
datasets produced similar results across various prune block sizes, with
comparable layers identified for removal. Given the consistency of results, we
opted to use Redpajama as the calibration dataset in subsequent experiments due
to its representative performance and alignment with our goals for efficient
model pruning. Recent studies around the time of this submission have explored
the nuanced role of calibration data in pruning large language
models~\citep{ji2024bewarecalibrationdatapruning}. While our analysis focuses on
the practical selection of calibration datasets, a deeper investigation into
calibration dataset characteristics and their influence on pruning decisions
remains an open question for future work.

\begin{table}[t]
  \centering
  \caption{Ablation study on the choice of calibration dataset for computing the
  angular cosine distance block importance metric on Llama3.1-8B Instruct. For
  calibration, we use a subset of 128 samples at a maximum sequence length (MSL)
  of 4096 from each dataset. The datasets C4, Redpajama, and Slimpajama produced
  similar pruning results across various block sizes, with comparable layers
  removed. Based on these results, the Redpajama dataset was selected for
  further evaluations due to its representative performance.}
  \begin{adjustbox}{max width=\textwidth}
  \begin{tabular}{cccc}
      \toprule
      Block Size & \begin{tabular}[c]{@{}c@{}}Removed\\
        Layers\end{tabular} & Dataset & \begin{tabular}[c]{@{}c@{}}Score\\
      (avg dist)\end{tabular} \\
      \midrule
      \multirow{3}{*}{2} & 24-25 & C4 & 0.145 \\
        & 23-24 & Redpajama & 0.168 \\
        & 24-25 & Slimpajama & 0.153 \\
      \midrule
      \multirow{3}{*}{4} & 24-27 & C4 & 0.197 \\
        & 23-26 & Redpajama & 0.222 \\
        & 23-26 & Slimpajama & 0.205 \\
      \midrule
      \multirow{3}{*}{6} & 22-27 & C4 & 0.241 \\
        & 22-27 & Redpajama & 0.270 \\
        & 23-28 & Slimpajama & 0.249 \\
        \midrule
        \multirow{3}{*}{8} & 20-27 & C4 & 0.282 \\
        & 20-27 & Redpajama & 0.293 \\
        & 20-27 & Slimpajama & 0.289 \\
      \bottomrule
  \end{tabular}
  \end{adjustbox}
  \label{tab:calibration_ablation_grouped_blocksize}
\end{table}

\subsection{Fine-tuning Datasets}
The following datasets were used for ablation studies and fine-tuning
experiments, representing a range of open-domain conversation,
instruction-following, reasoning, and mathematical tasks:

\begin{itemize}
    \item \textbf{Dolly 15k}~\citep{DatabricksBlog2023DollyV2} The Dolly dataset
    is an open-source collection of 15,000 instruction-following records
    generated by thousands of Databricks employees. It covers a wide range of
    behavioral categories, as outlined in
    InstructGPT\citep{ouyang2022traininglanguagemodelsfollow}, including
    brainstorming, classification, closed question answering (QA), generation,
    information extraction, open QA, and summarization. Dolly is designed to
    provide a benchmark for general-purpose instruction-following models,
    emphasizing diverse task types and behavioral categories.
    \item \textbf{GSM8k}~\citep{cobbe2021trainingverifierssolvemath} The GSM8k
    dataset is a collection of 8,000 high-quality grade-school-level math word
    problems, developed by OpenAI. Each problem is designed to assess a model's
    ability to perform multi-step reasoning and problem-solving, making it an
    essential benchmark for evaluating arithmetic, algebraic, and logical
    reasoning abilities in large models. Fine-tuning on GSM8k highlights the
    model's capacity for mathematical reasoning, a key focus of our ablation
    studies.
    \item \textbf{Alpaca
    Cleaned}\footnote{\url{https://huggingface.co/datasets/yahma/alpaca-cleaned}}~\citep{alpaca}
    The Alpaca Cleaned dataset is a cleaned version of the original Stanford
    Alpaca dataset, containing 51,760 instruction-following examples. It
    addresses several issues present in the original release, such as
    hallucinations, incorrect instructions, and output inconsistencies. This
    dataset provides high-quality general instruction-following tasks, spanning
    text generation, summarization, reasoning, and more. The cleaned version
    offers improved consistency and accuracy, making it ideal for fine-tuning
    large models in real-world instruction-following tasks.
    \item
    \textbf{OpenMathInstruct}~\citep{toshniwal2024openmath} The
    OpenMathInstruct-1 dataset is specifically designed for fine-tuning language
    models on mathematical instruction tasks. It contains 1.8 million
    problem-solution pairs, generated using
    Mixtral-8x7B~\citep{jiang2024mixtralexperts}. The problem sets are drawn
    from well-established mathematical benchmarks, including the GSM8K and
    MATH~\citep{hendrycksmath2021} datasets, ensuring a diverse and challenging
    range of mathematical reasoning tasks. Solutions are generated synthetically
    by allowing the Mixtral model to leverage a combination of natural language
    reasoning and executable Python code, which allows for both symbolic
    computation and procedural solutions. This combination of text and code
    execution makes the dataset particularly suited for training models to
    handle complex reasoning, problem-solving, and algebraic tasks.
\end{itemize}

\subsubsection{Data Sampling and Experimental Consistency}

To maintain consistency across ablation studies, we fixed the dataset size at
8,000 samples for GSM8k, Alpaca, and OpenMathInstruct, aligning them with the
standard GSM8k dataset size. However, the Dolly dataset retained its default
size of 15,000 samples to preserve the integrity of this benchmark. To evaluate
the impact of dataset size on self-data distillation, we extended the sample
sizes for some experiments, using the full 50,000 samples from Alpaca Cleaned
and randomly sampling 50,000 training samples from OpenMathInstruct. This
allowed us to control for the effects of larger datasets, providing insights
into how dataset size influences generalization and model retention following
pruning.

\subsection{Fine-tuning Pruned Models}

For fine-tuning, we employed Low-Rank Adaptation (LoRA)~\citep{hu2022lora}, as
it provides an efficient approach to training while preserving the pretrained
model's capacity. Although full fine-tuning is feasible, we focused on LoRA
fine-tuning in this study, leaving full parameter fine-tuning for future work.
We conducted a comprehensive grid search on an 8k-sample version of the
OpenMathInstruct dataset to identify the most effective hyperparameters for
LoRA-based fine-tuning. The search was performed across a range of values to
ensure optimal performance. We explored different \textit{rank sizes} $\in$ \{4,
8, 16, 32\}, aiming to balance model capacity and parameter efficiency. For the
\textit{number of epochs}, we tested values ranging $\in$ \{3, 5, 7, 10\},
ensuring that the models were fine-tuned enough to converge without overfitting.
The \textit{learning rate} was swept across five values \{2$\times$10$^{-5}$,
4$\times$10$^{-5}$, 6$\times$10$^{-5}$, 8$\times$10$^{-5}$,
1$\times$10$^{-4}$\}. Finally, we tested \textit{batch sizes} $\in$ \{8, 16, 32,
64, 128\} to determine the optimal balance between training stability and
computational efficiency.

Through this grid search, the optimal configuration was identified as a
\textit{rank size} = 8, \textit{epochs} = 5, a \textit{batch size} = 64, and
\textit{learning rate} = 1$\times$10$^{-4}$. These hyperparameters were used
consistently across all fine-tuning experiments (i.e., both standard supervised
fine-tuning and self-data distilled fine-tuning) in this study to ensure a fair
comparison of the models and their quality post-pruning. We conduct our model
training using \texttt{LLaMA-Factory
v0.8.3}\footnote{\url{https://github.com/hiyouga/LLaMA-Factory/releases/tag/v0.8.3}},
a versatile framework designed for large-scale language model training and
fine-tuning. This version offers extensive support for efficient parallelism,
optimized memory usage, and integration with popular datasets, making it ideal
for large model fine-tuning tasks such as those performed in this study.

\subsubsection{Computational Resources}

Fine-tuning and evaluations were conducted on Nvidia H100 GPUs. For experiments
involving larger self-data distillation datasets, we utilized Cerebras CS-3
Inference~\citep{cerebras2024inferencequality}, which achieves output generation
speeds exceeding 1800 tokens per second. The CS-3 system was particularly useful
for generating large-scale self-distilled datasets. However, for smaller
datasets (e.g., up to 15k samples), the H100 GPUs were sufficient for both
fine-tuning and generation.

\section{Extended Results on Fine-tuning Ablations}
\label{app:ablations_extended}
In this section, we provide extended results from our fine-tuning ablation study
to further clarify the impact of dataset choice on self-data distillation
efficacy in pruned Llama3.1-8B Instruct models. As detailed in the
Section~\ref{sec:ablations}, we observed that self-data distillation
consistently outperformed SFT across various datasets.
Table~\ref{tab:pruning_results_extended} shows that the largest gains were
achieved using the 50k-sample OpenMathInstruct dataset, particularly at medium
and large pruning block sizes (e.g., block size 6). At this configuration,
self-data distillation was able to recover 95.96\% of the baseline model
quality, which is a significant improvement compared to other datasets and
fine-tuning methods. This result highlights the robustness of the self-data
distillation process, especially in recovering quality post-pruning on
reasoning-heavy tasks like those in GSM8k, ARC-C, and MMLU.

Moreover, the recovery rates exhibited a clear trend where, larger datasets such
as the 50k OpenMathInstruct consistently led to higher quality retention,
especially when combined with more aggressive pruning. This suggests that the
dataset's ability to approximate the model's original data distribution is
critical for maintaining generalization capabilities after pruning. In contrast,
smaller datasets like Alpaca or Dolly showed comparatively lower recovery rates,
which further confirms the importance of dataset scale in the distillation
process. Our results suggest that larger datasets are crucial for mitigating
quality degradation in pruned models, with the 50k OpenMathInstruct dataset
emerging as the most effective in retaining and enhancing model quality across
block sizes, particularly in challenging reasoning tasks.

\section{Experimental Setup for Understanding Catastrophic Forgetting}
\label{app:catastrophic_forgetting}
To understand the impact of distribution shift on catastrophic forgetting, we
conducted experiments using the baseline model (i.e., Llama3.1-8B Instruct) and
its pruned variants fine-tuned with both supervised fine-tuning (SFT) and
self-data distilled fine-tuning (Self-Data FT). Specifically, we pruned 6
decoder layers, reducing the model from 32 to 26 layers, and evaluated the
models on the GSM8k dataset. For these experiments, we generated model responses
using the baseline and pruned variants on the GSM8k dataset to capture how the
distribution shift affects reasoning tasks post-pruning.
Following~\citet{Yang2024SelfDistillationBD}, to quantify the distribution
shift, we employed Sentence-BERT~\citep{Reimers2019SentenceBERTSE} to derive
sentence embeddings from the model-generated responses. Then, similar to the
method proposed by~\citet{zhang2023automatic}, we calculated the cosine
similarity between the sentence embeddings of the pruned models and those
generated by the original Llama3.1-8B Instruct model.

A lower cosine similarity score indicates a greater distribution shift,
suggesting a higher risk of catastrophic forgetting. Conversely, higher
similarity scores indicate better preservation of the original model's knowledge
and a lower risk of forgetting. These metrics allowed us to assess the extent to
which SFT and Self-Data FT preserved the learned distribution of the base model,
with the latter showing superior performance in mitigating forgetting, as
detailed in our ablations in Section~\ref{sec:ablations}.

\section{Model Merging Self-Data Distilled Models}
\label{app:model_merging}

We employ the Spherical Linear Interpolation (SLERP) method for merging pruned
models, which ensures smooth, geometrically consistent interpolation between two
pruned model parameter vectors. SLERP operates within the unit sphere's
geometry, contrasting with traditional linear interpolation that may destabilize
or yield suboptimal parameter combinations by ignoring the geometric properties
of the high-dimensional parameter space. SLERP preserves model integrity during
interpolation, leading to more stable and consistent outcomes.

Given two pruned model parameter vectors, $\boldsymbol{\theta}'_0$ and
$\boldsymbol{\theta}'_1$, corresponding to pruned models $M'_0$ (fine-tuned on
OpenMathInstruct) and $M'_1$ (fine-tuned on Alpaca), SLERP generates an
interpolated parameter vector $\boldsymbol{\theta}'_t$ for any interpolation
factor $t \in [0, 1]$. When $t = 0$, the parameters of the OpenMathInstruct
fine-tuned model $\boldsymbol{\theta}'_0$ are retrieved, and when $t = 1$, the
parameters of the Alpaca fine-tuned model $\boldsymbol{\theta}'_1$ are
retrieved.

\paragraph{Normalization to Unit Sphere}
The first step in SLERP is to normalize both pruned model parameter vectors to
lie on the unit sphere,
\[
\hat{\boldsymbol{\theta}}'_0 = \frac{\boldsymbol{\theta}'_0}{\|\boldsymbol{\theta}'_0\|}, \quad \hat{\boldsymbol{\theta}}'_1 = \frac{\boldsymbol{\theta}'_1}{\|\boldsymbol{\theta}'_1\|}.
\]
This normalization ensures that both parameter vectors have unit norms, placing
them on the surface of the unit sphere in the parameter space. Next, we compute
the angle $\theta_{\text{angle}}$ between the normalized pruned model vectors
$\hat{\boldsymbol{\theta}}'_0$ and $\hat{\boldsymbol{\theta}}'_1$. This angle is
computed using the dot product, $\cos(\theta_{\text{angle}}) =
\hat{\boldsymbol{\theta}}'_0 \cdot \hat{\boldsymbol{\theta}}'_1$, and the actual
angle is given by $\theta_{\text{angle}} =
\arccos(\cos(\theta_{\text{angle}}))$. This angle represents the angular
separation between the two pruned models' parameter vectors on the unit sphere.

\paragraph{Spherical Interpolation}
With the angle $\theta_{\text{angle}}$ determined, SLERP performs spherical
interpolation along the great circle connecting $\hat{\boldsymbol{\theta}}'_0$
and $\hat{\boldsymbol{\theta}}'_1$. The interpolated parameter vector
$\boldsymbol{\theta}'_t$ is computed as,
\[
\boldsymbol{\theta}'_t = \frac{\sin((1-t)\theta_{\text{angle}})}{\sin(\theta_{\text{angle}})} \cdot \hat{\boldsymbol{\theta}}'_0 + \frac{\sin(t\theta_{\text{angle}})}{\sin(\theta_{\text{angle}})} \cdot \hat{\boldsymbol{\theta}}'_1.
\]
This formula ensures that the interpolation remains on the surface of the unit
sphere, respecting the geometric structure of the parameter space. The
interpolation factor $t$ controls the contribution from each pruned model, when
$t = 0$, $\boldsymbol{\theta}'_t = \hat{\boldsymbol{\theta}}'_0$ (i.e.,
OpenMathInstruct fine-tuned model), and when $t = 1$, $\boldsymbol{\theta}'_t =
\hat{\boldsymbol{\theta}}'_1$ (i.e., Alpaca fine-tuned model). The intermediate
values of $t$ produce a smooth, spherical blend of the two pruned models.

\subsection{Geometric Consistency and Application}
By operating within the unit sphere, SLERP respects the \textit{Riemannian
geometry} of the high-dimensional parameter space, ensuring a smooth transition
between the two pruned models. Traditional linear interpolation in such spaces
can distort the relationships between parameters, leading to suboptimal
combinations and degraded model performance. In contrast, SLERP maintains
geometric consistency, ensuring that the interpolation follows a natural path on
the unit sphere.

Merging the pruned OpenMathInstruct and Alpaca models using SLERP combines the
unique strengths of both models. For instance, OpenMathInstruct's emphasis on
mathematical reasoning and logical structure complements Alpaca's broader
instruction-following capabilities. By adjusting the interpolation factor $t$,
the merged model can balance these capabilities, resulting in a versatile and
robust model for a range of downstream tasks. We use Arcee.ai's
\texttt{mergekit}\footnote{\url{https://github.com/arcee-ai/mergekit}} for
efficiently merging model checkpoints.

\begin{table*}
    \vspace{-5pt}
    \caption{\textbf{Model quality results for pruned Llama3.1-8B Instruct models across various pruning block sizes and fine-tuning strategies.} 
    This table reports the quality of different fine-tuning methods (No
    Fine-tuning, Standard Fine-tuning (SFT), and Self-Data Distillation) on
    various datasets, with average accuracy across ARC-C, GSM8k, and MMLU tasks.
    The "Avg. Recovery" column shows the percentage of model quality recovered
    relative to the unpruned baseline. The table highlights that the self-data
    distillation strategy consistently yields superior recovery rates,
    particularly with the 50k-sample OpenMathInstruct dataset. For instance, at
    a pruning block size of 6, the self-data distilled OpenMathInstruct model
    retains 95.96\% of the original unpruned Llama3.1-8B Instruct (i.e., 32
    layers) model's quality, the highest recovery observed among all datasets
    and fine-tuning methods.}

    \label{tab:pruning_results_extended}
    \centering
    \begin{adjustbox}{max width=\textwidth}
    \begin{tabular}{ccccccccccc}
        \toprule
        \begin{tabular}[c]{@{}c@{}}Prune\\ Block Size\end{tabular}&
        \begin{tabular}[c]{@{}c@{}}Model\\ Savings\end{tabular} &
        \begin{tabular}[c]{@{}c@{}}Fine-tuning\\ Method\end{tabular} & Dataset &
        \begin{tabular}[c]{@{}c@{}}ARC-C\\ (25-shot)\end{tabular} &
        \begin{tabular}[c]{@{}c@{}}GSM8k\\ (5-shot)\end{tabular}  &
        \begin{tabular}[c]{@{}c@{}}MMLU\\ (5-shot)\end{tabular}  &
        \begin{tabular}[c]{@{}c@{}}Avg.\\ Score\end{tabular}  &
        \begin{tabular}[c]{@{}c@{}}Avg.\\ Recovery \end{tabular} \\
        \midrule
        Baseline &  -    & \multicolumn{1}{l}{No FT}  &   & 58.70 & 63.15 &
        67.40 & 63.08 & 100.00\% \\
        \midrule
        \multirow{10}{*}{2} &
          \multirow{10}{*}{\begin{tabular}[c]{@{}c@{}}5.43\%\\ (7.59B)
          \end{tabular}} & \multicolumn{1}{l}{No FT} & & 55.20 & 67.79 & 56.18 &
          59.72 & 94.67\% \\
        &  & \multicolumn{1}{l}{SFT}  & \multicolumn{1}{l}{GSM8k}  &
        \textbf{58.45} & 56.25 & 65.22 & 59.97 & 95.07\% \\
        & & \multicolumn{1}{l}{Self-Data Distillation}  &
        \multicolumn{1}{l}{GSM8k}  & 57.34 & 64.44 & 66.60 & 62.79 & 99.54\% \\
        \cmidrule{3-4}
        &  & \multicolumn{1}{l}{SFT}  & \multicolumn{1}{l}{Dolly}  & 55.67 &
         61.64 & 65.71 & 61.01 & 96.71\% \\
        & & \multicolumn{1}{l}{Self-Data Distillation}  &
        \multicolumn{1}{l}{Dolly}  & 56.48 & 62.24 & 66.46 & 61.73 & 97.87\% \\
        \cmidrule{3-4}
        &  & \multicolumn{1}{l}{SFT}  & \multicolumn{1}{l}{Alpaca (50k)}  &
         56.61 & 63.19 & 65.60 & 61.80 & 97.98\% \\
        & & \multicolumn{1}{l}{Self-Data Distillation}  &
        \multicolumn{1}{l}{Alpaca (50k)}  & 56.91 & 68.60 & \textbf{66.50} &
        63.34 & 100.41\% \\
        \cmidrule{3-4}
        &  & \multicolumn{1}{l}{SFT}  & \multicolumn{1}{l}{OpenMathInstruct
         (50k)}  & 52.91 & 44.95 & 60.93 & 52.93 & 83.88\% \\
        & & \multicolumn{1}{l}{Self-Data Distillation}  &
        \multicolumn{1}{l}{OpenMathInstruct (50k)}  & 56.43  & \textbf{69.97} &
        66.17 & \textbf{64.19} & \textbf{101.76\%} \\
        \midrule
        \multirow{10}{*}{4} & \multirow{10}{*}{
          \begin{tabular}[c]{@{}c@{}}10.86\%\\ (7.16B) \end{tabular}} &
          \multicolumn{1}{l}{No FT} & & 55.20 & 56.18 & \textbf{67.79} & 59.72 &
          94.67\% \\
        &  & \multicolumn{1}{l}{SFT}  & \multicolumn{1}{l}{GSM8k}  & 54.27 &
        43.47 & 65.40 & 54.38 & 86.22\% \\
        & & \multicolumn{1}{l}{Self-Data Distillation}  &
        \multicolumn{1}{l}{GSM8k}  & 55.20 & 62.55 & 66.68 & 61.48 & 97.49\% \\
        \cmidrule{3-4}
        &  & \multicolumn{1}{l}{SFT}  & \multicolumn{1}{l}{Dolly}  & 54.78 &
         59.36 & 63.65 & 59.26 & 93.95\% \\
        & & \multicolumn{1}{l}{Self-Data Distillation}  &
        \multicolumn{1}{l}{Dolly}  & 51.71 & 55.96 & 65.40 & 57.69 & 91.44\% \\
        \cmidrule{3-4}
        &  & \multicolumn{1}{l}{SFT}  & \multicolumn{1}{l}{Alpaca (50k)}  &
        56.05 & 54.40 & 65.34 & 58.60 & 92.89\% \\
        & & \multicolumn{1}{l}{Self-Data Distillation}  &
        \multicolumn{1}{l}{Alpaca (50k)}  & \textbf{57.27} & 66.20 & 66.24 &
        \textbf{63.24} & \textbf{100.26\%} \\
        \cmidrule{3-4}
        &  & \multicolumn{1}{l}{SFT}  & \multicolumn{1}{l}{OpenMathInstruct
        (50k)}  & 51.62 & 44.35 & 61.65 & 52.54 & 83.30\% \\
        & & \multicolumn{1}{l}{Self-Data Distillation}  &
        \multicolumn{1}{l}{OpenMathInstruct (50k)}  &  53.93 & \textbf{69.44} &
        65.34 & 62.24 & 98.66\% \\
        \midrule
        \multirow{10}{*}{6} &
          \multirow{10}{*}{\begin{tabular}[c]{@{}c@{}}16.30\%\\ (6.72B)
          \end{tabular}} & \multicolumn{1}{l}{No FT} & & 49.49 & 0.00 &
          \textbf{67.42} & 48.93 & 70.50\% \\
        & & \multicolumn{1}{l}{SFT}  & \multicolumn{1}{l}{GSM8k}  & 46.67 &
        62.09 & 64.67 & 57.81 & 91.63\% \\
        &  & \multicolumn{1}{l}{Self-Data Distillation}  &
        \multicolumn{1}{l}{GSM8k}  & 51.45 & 60.05 & 66.28 & 59.93 & 95.02\% \\
        \cmidrule{3-4}
        &  & \multicolumn{1}{l}{SFT}  & \multicolumn{1}{l}{Dolly}  & 47.18 &
        33.89 & 65.33 & 48.13 & 76.31\% \\
        & & \multicolumn{1}{l}{Self-Data Distillation}  &
        \multicolumn{1}{l}{Dolly}  & 51.96 & 50.95 & 62.56 & 55.82 & 88.47\%  \\
         \cmidrule{3-4}
        &  & \multicolumn{1}{l}{SFT}  & \multicolumn{1}{l}{Alpaca (50k)}  &
        \textbf{54.62} & 56.11 & 64.39 & 58.37 & 92.53\% \\
        & & \multicolumn{1}{l}{Self-Data Distillation}  &
        \multicolumn{1}{l}{Alpaca (50k)}  & 53.80 & 59.15 & 66.29 & 59.75 &
        94.71\% \\
        \cmidrule{3-4}
        &  & \multicolumn{1}{l}{SFT}  & \multicolumn{1}{l}{OpenMathInstruct
         (50k)}  & 46.93 & 43.82 & 59.98 & 50.91 & 80.71\% \\
        & & \multicolumn{1}{l}{Self-Data Distillation}  &
        \multicolumn{1}{l}{OpenMathInstruct (50k)}  & 50.00 & \textbf{66.64} &
        64.96 & \textbf{60.53} & \textbf{95.96\%} \\
        \midrule
        \multirow{10}{*}{8} &
          \multirow{10}{*}{\begin{tabular}[c]{@{}c@{}}21.73\%\\ (6.29B)
          \end{tabular}} & \multicolumn{1}{l}{No FT} & & 44.71 & 0.00 &
          \textbf{65.57} & 36.76 & 58.27\% \\
        &  & \multicolumn{1}{l}{SFT}  & \multicolumn{1}{l}{GSM8k}  & 44.79 &
        50.86 & 64.38 & 53.34 & 84.56\% \\
        &  & \multicolumn{1}{l}{Self-Data Distillation}  &
        \multicolumn{1}{l}{GSM8k}  & 46.16 & 50.11 & 65.53 & 53.93 & 85.50\% \\
        \cmidrule{3-4}
        &  & \multicolumn{1}{l}{SFT}  & \multicolumn{1}{l}{Dolly}  & 39.33 &
          15.39 & 57.44 & 37.39 & 59.27\% \\
        & & \multicolumn{1}{l}{Self-Data Distillation}  &
        \multicolumn{1}{l}{Dolly}  & 46.50 & 28.73 & 62.93 & 46.05 & 73.00\% \\
        \cmidrule{3-4}
        &  & \multicolumn{1}{l}{SFT}  & \multicolumn{1}{l}{Alpaca (50k)}  &
        \textbf{49.15} & 39.59 & 64.81 & 51.85 & 82.19\% \\
        & & \multicolumn{1}{l}{Self-Data Distillation}  &
        \multicolumn{1}{l}{Alpaca (50k)}  & 48.09 & 42.77 & 65.20 & 52.69 &
        83.51\% \\
        \cmidrule{3-4}
        &  & \multicolumn{1}{l}{SFT}  & \multicolumn{1}{l}{OpenMathInstruct
        (50k)}  & 42.15 & 29.64 & 60.65 & 44.81 & 71.05\% \\
        & & \multicolumn{1}{l}{Self-Data Distillation}  &
        \multicolumn{1}{l}{OpenMathInstruct (50k)}  & 46.67 & \textbf{57.70} &
        64.87 & \textbf{56.41} & \textbf{89.44\%} \\
        \midrule
        \multirow{10}{*}{10} &
          \multirow{10}{*}{\begin{tabular}[c]{@{}c@{}}27.16\%\\ (5.85B)
          \end{tabular}} & \multicolumn{1}{l}{No FT} & & 37.46 & 0.00 & 64.09 &
          33.85 & 53.65\% \\
        &  & \multicolumn{1}{l}{SFT}  & \multicolumn{1}{l}{GSM8k}  & 39.85 &
        37.45 & 61.47 & 46.92 & 74.36\% \\
        &  & \multicolumn{1}{l}{Self-Data Distillation}  &
        \multicolumn{1}{l}{GSM8k}  & 41.55 & 37.47 & 62.33 & 47.12 & 74.70\% \\
        \cmidrule{3-4}
        &  & \multicolumn{1}{l}{SFT}  & \multicolumn{1}{l}{Dolly}  & 38.40 &
        0.76 & 46.94 & 28.70 & 45.51\% \\
        & & \multicolumn{1}{l}{Self-Data Distillation}  &
        \multicolumn{1}{l}{Dolly}  & 43.60 & 12.28 & 62.47 & 39.45 & 62.53\% \\
        \cmidrule{3-4}
        &  & \multicolumn{1}{l}{SFT}  & \multicolumn{1}{l}{Alpaca (50k)}  &
         \textbf{45.22} & 17.51 & 63.72 & 42.82 & 67.88\% \\
        & & \multicolumn{1}{l}{Self-Data Distillation}  &
        \multicolumn{1}{l}{Alpaca (50k)}  & 44.91 & 20.05 & \textbf{64.73} &
        43.90 & 69.56\% \\
        \cmidrule{3-4}
        &  & \multicolumn{1}{l}{SFT}  & \multicolumn{1}{l}{OpenMathInstruct
        (50k)}  & 39.33 & 15.39 & 57.44 & 37.39 & 59.25\% \\
        & & \multicolumn{1}{l}{Self-Data Distillation}  &
        \multicolumn{1}{l}{OpenMathInstruct (50k)}  & 40.88  & \textbf{44.58} &
        64.54 & \textbf{50.33} & \textbf{79.79}\% \\
        \bottomrule
    \end{tabular}
    \end{adjustbox}
    \vspace{-5pt}
\end{table*}